\documentclass[lettersize,journal]{IEEEtran}
\usepackage{amsmath,amsfonts}
\usepackage{algorithmic}
\usepackage{algorithm}
\usepackage{array}
\usepackage[caption=false,font=normalsize,labelfont=sf,textfont=sf]{subfig}
\usepackage{textcomp}
\usepackage{stfloats}
\usepackage{url}
\usepackage{verbatim}
\usepackage{graphicx}
\usepackage{cite}
\usepackage{booktabs}

\usepackage{amsmath,amsfonts,bm}

\def\eqref#1{equation~\ref{#1}}

\def\1{\bm{1}}

\def\vx{{\bm{x}}}
\def\vy{{\bm{y}}}

\def\mA{{\bm{A}}}
\def\mB{{\bm{B}}}

\def\mD{{\bm{D}}}

\def\mH{{\bm{H}}}
\def\mI{{\bm{I}}}

\def\mL{{\bm{L}}}
\def\mM{{\bm{M}}}
\def\mN{{\bm{N}}}

\def\mS{{\bm{S}}}

\def\mX{{\bm{X}}}

\DeclareMathAlphabet{\mathsfit}{\encodingdefault}{\sfdefault}{m}{sl}
\SetMathAlphabet{\mathsfit}{bold}{\encodingdefault}{\sfdefault}{bx}{n}

\def\gE{{\mathcal{E}}}

\def\gG{{\mathcal{G}}}

\def\gN{{\mathcal{N}}}

\def\gV{{\mathcal{V}}}

\usepackage{multicol}
\usepackage{multirow}
\usepackage{wrapfig}
\usepackage{hyperref}
\usepackage{cleveref}

\usepackage{xspace}
\newcommand{\etal}{\textit{et al.}\xspace}

\begin{document}

\title{Hyper-YOLO: When Visual Object Detection \\ Meets Hypergraph Computation}

\author{
Yifan~Feng, 
Jiangang~Huang,
Shaoyi~Du,~\IEEEmembership{Senior~Member,~IEEE,}
Shihui~Ying, 
Jun-Hai~Yong,
Yipeng~Li,
Guiguang~Ding,
Rongrong~Ji,~\IEEEmembership{Senior~Member,~IEEE}
Yue~Gao,~\IEEEmembership{Senior~Member,~IEEE}
\thanks{Yifan Feng, Jun-Hai Yong, Guiguang Ding, and Yue Gao are with the School of Software, BNRist, THUIBCS, BLBCI, Tsinghua University, Beijing 100084, China. 
E-mail: evanfeng97@gmail.com; \{yongjh,dinggg, gaoyue\}@tsinghua.edu.cn;}
\thanks{Jiangang Huang and Shaoyi Du are with Institute of Artificial Intelligence and Robotics, College of Artificial Intelligence, Xi’an Jiaotong University, Xi’an 710049, China. 
E-mail: mywhy666@stu.xjtu.edu.cn; dushaoyi@xjtu.edu.cn;}
\thanks{Shihui Ying is with the Department of Mathematics, School of Science, Shanghai University, Shanghai 200444, China. 
E-mail: shying@shu.edu.cn;}
\thanks{Yipeng Li is with the Department of Automation, Tsinghua University, Beijing 100084, China. 
E-mail: liyipg@gmail.com;}
\thanks{Rongrong Ji is with the Media Analytics and Computing Laboratory, Department of Artificial Intelligence, School of Informatics, Institute of Artificial Intelligence, and Fujian Engineering Research Center of Trusted Artificial Intelligence Analysis and Application, Xiamen University, 361005, China.
E-mail: rrji@xmu.edu.cn;
}
}

\markboth{Journal of \LaTeX\ Class Files,~Vol.~14, No.~8, August~2021}%
{Shell \MakeLowercase{\textit{et al.}}: A Sample Article Using IEEEtran.cls for IEEE Journals}


\maketitle

\begin{abstract}
We introduce Hyper-YOLO, a new object detection method that integrates hypergraph computations to capture the complex high-order correlations among visual features. Traditional YOLO models, while powerful, have limitations in their neck designs that restrict the integration of cross-level features and the exploitation of high-order feature interrelationships. To address these challenges, we propose the Hypergraph Computation Empowered Semantic Collecting and Scattering (HGC-SCS) framework, which transposes visual feature maps into a semantic space and constructs a hypergraph for high-order message propagation. This enables the model to acquire both semantic and structural information, advancing beyond conventional feature-focused learning.
Hyper-YOLO incorporates the proposed Mixed Aggregation Network (MANet) in its backbone for enhanced feature extraction and introduces the Hypergraph-Based Cross-Level and Cross-Position Representation Network (HyperC2Net) in its neck. HyperC2Net operates across five scales and breaks free from traditional grid structures, allowing for sophisticated high-order interactions across levels and positions. This synergy of components positions Hyper-YOLO as a state-of-the-art architecture in various scale models, as evidenced by its superior performance on the COCO dataset. Specifically, Hyper-YOLO-N significantly outperforms the advanced YOLOv8-N and YOLOv9-T with 12\% $\text{AP}^{val}$ and 9\% $\text{AP}^{val}$ improvements. The source codes are at \href{https://github.com/iMoonLab/Hyper-YOLO}{https://github.com/iMoonLab/Hyper-YOLO}.
\end{abstract}


\section{Introduction}
\label{sec:intro}

The YOLO series \cite{yolo, yolov3, yolov4, yolov5, yolov6, yolov6-init, yolov7, yolov8, yolox, goldyolo, pp-yoloe} stands out as a mainstream method in the realm of object detection, offering several advantages that cater to these diverse applications. 
The architecture of YOLO consists of two main components: the backbone\cite{efficientnet, yolov7, ding2021repvgg, ELAN} and neck\cite{FPN, PANet, goldyolo}. While the backbone is designed for feature extraction and has been extensively studied, the neck is responsible for the fusion of multi-scale features, providing a robust foundation for the detection of variously sized objects. This paper places a particular emphasis on the neck, which is paramount in enhancing the model's ability to detect objects across different scales.

\begin{figure}
    \centering
    \includegraphics[width=0.48\textwidth]{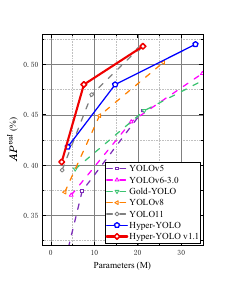}
    \caption{Comparison with other SOTA YOLO series methods on the COCO.}
    \label{fig:curve}
\end{figure}

Contemporary YOLO models have adopted the PANet\cite{PANet} for their necks, which employs top-down and bottom-up pathways to facilitate a comprehensive fusion of information across scales. However, the PANet's capability is predominantly confined to fusing features between adjacent layers and does not sufficiently address cross-level feature integration. In contrast, the gather-distribute neck design, exemplified by Gold-YOLO\cite{goldyolo}, promotes inter-layer information exchange but still falls short of facilitating cross-position interactions within the feature map. Moreover, it does not thoroughly explore the potential of feature interrelationships, particularly those involving high-order correlations. High-order correlations refer to the complex and often non-linear relationships that exist between features at different scales, positions, and semantic levels, which are critical for understanding the deeper context and interactions within visual data.
It is noticed that the synergistic representation of low-level visual features and their correlations plays a critical role in the object detection task. The integration of these basic features with high-level semantic information is crucial for the accurate identification and localization of objects within a given scene.
The exploration of high-order correlations underlying low-level features for semantic analysis remains a challenging yet essential topic within many computer vision tasks. This phenomenon, where the mining of such high-order relationships is commonly overlooked, may limit the performance of vision tasks.

In practice, hypergraphs \cite{gao2020hypergraph, HGNNP} are commonly employed to represent complex high-order correlations due to their enhanced expressive power over simple graphs. While edges in a simple graph are limited to connecting only two vertices, thereby greatly restricting their expressiveness, hyperedges in a hypergraph can connect two or more vertices, enabling the modeling of more intricate high-order relationships.
Compared to simple graphs, hypergraphs can capture a richer set of interactions among multiple entities, which is vital for tasks that require an understanding of complex and multi-way relationships, such as object detection in computer vision, where those cross-level and cross-position correlations among feature maps are crucial.

Different from most previous works focusing on enhancing the backbone of feature extraction, we propose the Hypergraph Computation Empowered Semantic Collecting and Scattering (HGC-SCS) framework. This framework is ingeniously conceived to enhance the feature maps extracted by the visual backbone through their transposition into an abstract semantic space, followed by the construction of an intricate hypergraph structure. The hypergraph serves as a conduit for enabling high-order message propagation among the features within this semantic space. Such an approach equips the visual backbone with the dual capability of assimilating both \textbf{semantic} and complex \textbf{structural information}, thereby overcoming the limitations of conventional semantic feature-focused learning and elevating performance beyond its traditional bounds.

Building upon the aforementioned HGC-SCS framework, we introduce Hyper-YOLO, a new YOLO method based on hypergraph computation. Hyper-YOLO, for the first time, integrates hypergraph computation within the neck component of a visual target detection network. By modeling the intricate high-order associations inherent to feature maps extracted from the backbone, Hyper-YOLO substantially enhances object detection performance. 
In terms of the backbone architecture, Hyper-YOLO incorporates the Mixed Aggregation Network (MANet), which amalgamates three distinctive foundational structures to enrich the flow of information and augment feature extraction capabilities, building upon the base provided by YOLOv8. In the realm of the neck, leveraging the proposed HGC-SCS framework, we achieve a multi-scale feature fusion neck known as the Hypergraph-Based Cross-Level and Cross-Position Representation Network (HyperC2Net). In contrast to conventional neck designs, HyperC2Net fuses features across five different scales, concurrently breaking away from the grid structure of visual feature maps to facilitate high-order message propagation across \textbf{levels} and \textbf{positions}.
The combined enhancements in both the backbone and the neck position Hyper-YOLO as a groundbreaking architecture. The empirical results (\Cref{fig:curve}) on the COCO dataset attest to its significant superiority in performance, substantiating the efficacy of this sophisticated approach in advancing the field of object detection.
Our contributions can be summarized as:
\begin{enumerate}
    \item We propose the Hypergraph Computation Empowered Semantic Collecting and Scattering (HGC-SCS) framework, enhancing visual backbones with high-order information modeling and learning. 
    \item Leveraging the proposed HGC-SCS framework, we develop HyperC2Net, an object detection neck that facilitates high-order message passing throughout semantic layers and positions. HyperC2Net markedly elevates the neck's proficiency in distilling high-order features.
    \item We propose the Mixed Aggregation Network (MANet), which incorporates three types of blocks to enrich the information flow, thereby enhancing the feature extraction capabilities of the backbone. 
    \item We present Hyper-YOLO, which incorporates hypergraph computations to enhance the model's high-order information perception capabilities, leading to improvements in object detection.
    Specifically, our Hyper-YOLO-N achieves significant improvements in $\text{AP}^{val}$, with a 12\% increase compared to YOLOv8-N and a 9\% increase compared to YOLOv9-T on the COCO dataset.
\end{enumerate}

\section{Related Work}
\label{sec:related}

\subsection{YOLO Series Object Detectors}
The YOLO series has been a cornerstone in real-time object detection, evolving from YOLOv1's \cite{yolo} single-stage detection to YOLOv8's \cite{yolov8} performance-optimized models. Each iteration, from YOLOv4's \cite{yolov4} structural refinements to YOLOv7's \cite{yolov7} E-ELAN backbone, has brought significant advancements. YOLOX \cite{yolox} introduced anchor-free detection, and Gold-YOLO \cite{goldyolo} enhanced feature fusion with its Gather-and-Distribute mechanism. Despite the emergence of RT-DETR \cite{RTMDet} and other detectors, the YOLO series remains prevalent, partly due to its effective use of CSPNet, ELAN \cite{ELAN}, and improved PANet \cite{PANet} or FPN \cite{FPN} for feature integration, coupled with sophisticated prediction heads from YOLOv3 \cite{yolov3} and FCOS \cite{FCOS}. YOLOv9\cite{yolov9} introduces programmable gradient information and the Generalized Efficient Layer Aggregation Network to minimize information loss during deep network transmission. Building upon those YOLO methods, this paper presents Hyper-YOLO, an advanced approach that leverages hypergraph computations to enhance the complex correlation learning capabilities of the YOLO framework. Hyper-YOLO aims to improve the learning and integration of hierarchical features, pushing the boundaries of object detection performance.

\subsection{Hypergraph Learning Methods}
A hypergraph \cite{gao2020hypergraph, HGNNP} can be utilized to capture these complex, high-order associations. Hypergraphs, with their hyperedges connecting multiple nodes, excel in modeling intricate relationships, as evidenced in their application to diverse domains such as social network analysis\cite{young2021hypergraph, yang2020lbsn2vec++}, drug-target interaction modeling\cite{jin2023general, vinas2023hypergraph}, and brain network analysis \cite{xiao2019multi, zu2016identifying}. 
Hypergraph learning methods have emerged as a powerful tool for capturing complex and high-order correlations in data, which traditional graph-based techniques may not adequately represent. The notion of hyperedges, as discussed in Gao \etal \cite{gao2020hypergraph}, facilitates the modeling of these intricate relationships by allowing multiple nodes to interact simultaneously. Hypergraph Neural Networks (HGNN) \cite{HGNN} exploit these relationships, enabling direct learning from hypergraph structures through spectral methods. Building on this, General Hypergraph Neural Networks (HGNN$^+$) \cite{HGNNP} introduce spatial approaches for high-order message propagation among vertices, further expanding the capabilities of hypergraph learning.
Despite these advancements, the application of hypergraph learning in computer vision tasks remains relatively unexplored, particularly in the areas of modeling and learning high-order associations. In this paper, we will delve into how hypergraph computations can be harnessed for object detection tasks, aiming to elevate both classification and localization accuracy by integrating the nuanced relational information that modeled by the hypergraph.

\section{Hypergraph Computation Empowered Semantic Collecting and Scattering Framework}

Unlike representation learning in computer vision only processes visual features, those hypergraph computation methods \cite{HGNN, HGNNP} simultaneously process features and high-order structures. Most hypergraph computation methods rely on the inherent hypergraph structures, which cannot be obtained in most computer vision scenarios. 
Here, we introduce the general paradigm of hypergraph computation in computer vision, which includes hypergraph construction and hypergraph convolution. Given the feature map $\mX$ extracted from the neural networks, the hypergraph construction function $f: \mX \rightarrow \gG$ is adopted to estimate the potential high-order correlations among feature points in the semantic space. Then, the spectral or spatial hypergraph convolution methods are utilized to propagate high-order messages among feature points via the hypergraph structure. The generated high-order features are termed $\mX_{hyper}$. 
By integrating high-order relational information into $ \mathbf{X}_{hyper} $, this hypergraph computation strategy addresses the deficiency of high-order correlations in the original feature map $ \mathbf{X} $. 
The resultant hybrid feature map, denoted as $ \mathbf{X}' $, emerges from the fusion of $ \mathbf{X} $ and $ \mathbf{X}_{\text{hyper}} $. This synthesized process culminates in a semantically enhanced visual representation $ \mathbf{X}' $, which provides more comprehensive visual feature representations from both semantic and high-order structural perspectives.

Here, we devise a general framework for hypergraph computation in computer vision, named the Hypergraph Computation Empowered Semantic Collecting and Scattering (HGC-SCS) framework. Given those feature maps extracted from CNN \cite{he2016deep,he2016identity,huang2017densely,liu2022convnet,szegedy2016rethinking,xie2017aggregated} or other backbones, our framework first collects those features and fuses them to construct the mixed feature bag $\mX_{mixed}$ in the semantic space. In the second step, we estimate those potential high-order correlations to construct the hypergraph structure in the semantic space. To fully utilize those high-order structure information, some related hypergraph computation methods \cite{HGNN, HGNNP} can be employed. In this way, the \textbf{high-order aware feature} $\mX_{hyper}$ can be generated, which incorporates both the high-order structural and the semantic information. In the last step, we scatter the high-order structural information to each input feature map. The HGC-SCS framework can be formulated as follows:
\begin{equation}
\small
\nonumber
    \left\{ 
        \begin{aligned}
            & \mX_{mixed} \xleftarrow{\text{Collecting}} \{\mX_1, \mX_2, \cdots \} \\
            & \mX_{hyper} = \text{HyperComputation}(\mX_{mixed}) \quad\textit{//High-Order Learning} \\
            & \{\mX'_1, \mX'_2, \cdots \} \xleftarrow{\text{Scattering}} \{\phi(\mX_{hyper}, \mX_1), \phi(\mX_{hyper}, \mX_2), \cdots \}
        \end{aligned}
    \right. ,
\end{equation}
where $\{\mX_1, \mX_2, \cdots\}$ denotes the basic feature maps generated from the visual backbone. The ``HyperComputation'' denotes the second step, including hypergraph construction and hypergraph convolution, which captures those potential high-order structural information in the semantic space and generates the high-order aware feature $\mX_{hyper}$. In the last line, $\phi(\cdot)$ denotes the feature fusion function. $\{\mX'_1, \mX'_2, \cdots \}$ denotes the enhanced visual feature maps. In the following, we will introduce an instance of our HGC-SCS framework in object detection named HyperC2Net.

\section{Methods}
\label{sec:method}
In this section, we first introduce preliminaries of YOLO notations as well as the framework of the proposed Hyper-YOLO. 
In the following, we detail the proposed two core modules, including the basic block (MANet) and neck (HyperC2Net) of our Hyper-YOLO. Finally, we analyze the relationship between Hyper-YOLO and other YOLO methods.

\subsection{Preliminaries}
The YOLO series methods \cite{yolo, yolov3, yolov4, yolov5, yolov6, yolov7, yolov8,yolov9,YOLO-MS,DAMO-YOLO,Scaled-YOLOV4,YOLO9000,YOLOCS} are typically composed of two main components: backbone and neck. The backbone \cite{lee2019energy} \cite{ding2021repvgg} is responsible for extracting fundamental visual features, while the neck \cite{FPN} \cite{PANet} \cite{RTMDet} facilitates the fusion of multi-scale features for the final object detection. This paper proposes enhancement strategies specifically targeting these two components. For ease of description within this paper, we denote the three scale outputs of the neck as $\{\mN_3, \mN_4, \mN_5\}$, corresponding respectively to small-scale, medium-scale, and large-scale detection. In the feature extraction phase of the backbone, we further divide it into five stages: $\{\mB_1, \mB_2, \mB_3, \mB_4, \mB_5\}$, which represent features at different semantic levels. A larger number indicates that the feature is a higher-level semantic feature extracted by a deeper layer of the network. More details are provided in \cref{app:sec:details}.

\subsection{Hyper-YOLO Overview}

Our Hyper-YOLO framework maintains the overall architecture of the typical YOLO methods, including the backbone and neck, as depicted in \cref{app:fig:details}. Given an image, the backbone of Hyper-YOLO leverages the proposed MANet as its core computational module, thereby augmenting the feature discernment ability of the conventional C2f module found in YOLOv8 \cite{yolov8}. Diverging from traditional YOLO architectures, Hyper-YOLO ingests an ensemble of five primary feature sets $\{\mB_1, \mB_2, \mB_3, \mB_4, \mB_5\}$. In a novel stride, the neck (HyperC2Net) of Hyper-YOLO, grounded in hypergraph computational theory, integrates cross-level and cross-position information across these quintuple feature sets, culminating in the generation of final semantic features $\{\mN_3, \mN_4, \mN_5\}$ across three distinct scales. These hierarchically structured semantic features are subsequently harnessed for the final object detection task.

\begin{figure*}
    \centering
    \includegraphics[width=0.8\textwidth]{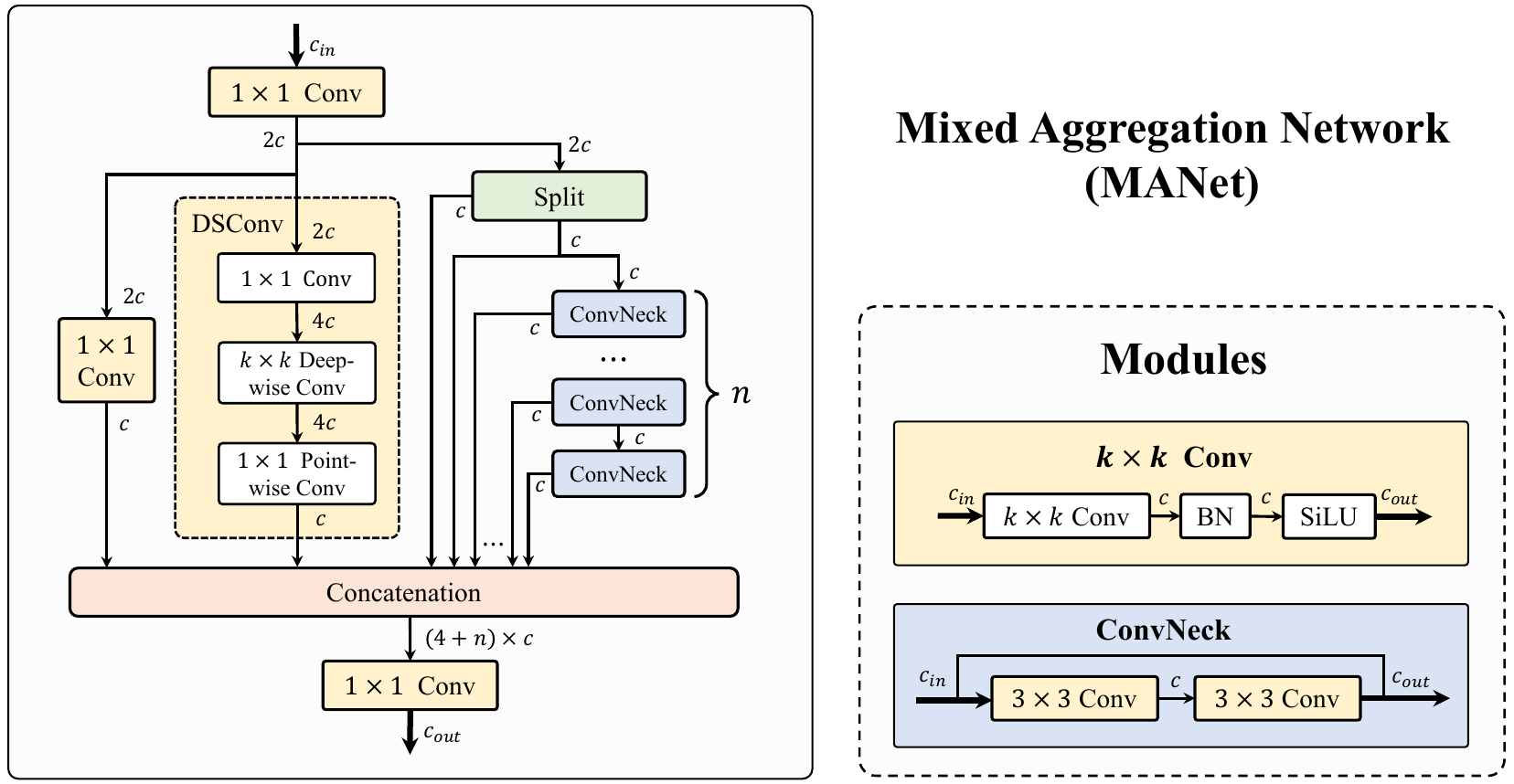}
    \caption{Illustration of the proposed Mixed Aggregation Network (MANet).}
    \label{fig:bb}
\end{figure*}

\begin{figure*}
    \centering
    \includegraphics[width=0.8\textwidth]{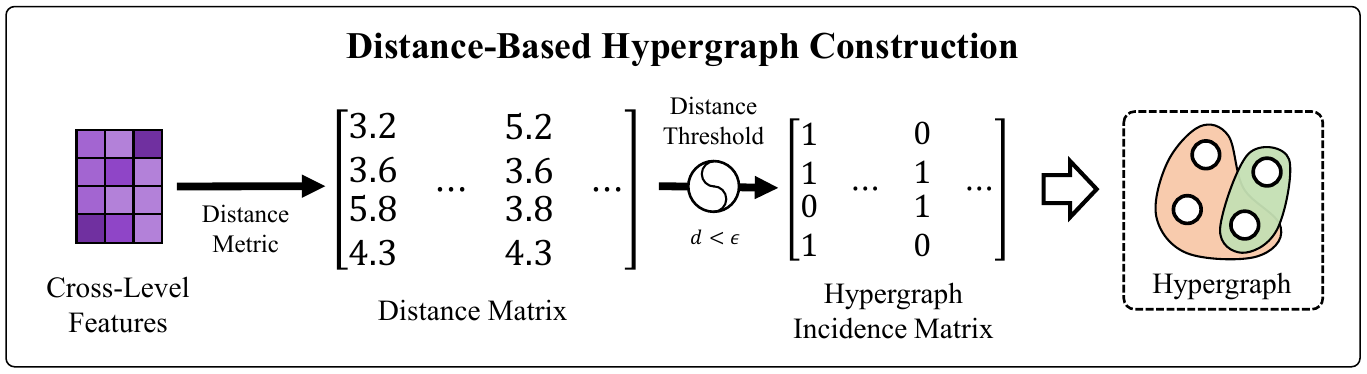}
    \caption{Illstration of hypergraph construction.}
    \label{fig:hg_con}
\end{figure*}

\subsection{Mixed Aggregation Network}
As for the backbone of our Hyper-YOLO, to augment the feature extraction prowess of the foundational network, we devise the Mixed Aggregation Network (MANet), as shown in \cref{fig:bb}. This architecture synergistically blends three typical convolutional variants: the $1 \times 1$ bypass convolution for channel-wise feature recalibration, the Depthwise Separable Convolution (DSConv) for efficient spatial feature processing, and the C2f module for enhancing feature hierarchy integration. This confluence produces a more variegated and rich gradient flow during the training phase, which significantly amplifies the semantic depth encapsulated within the base features at each of the five pivotal stages. Our MANet can be formulated as follows:
\begin{equation}
    \left\{ 
    \begin{aligned}
        &\mX_{mid} = \text{Conv}_1(\mX_{in}) \\ 
        &\mX_{1} = \text{Conv}_2(\mX_{mid}) \\ 
        &\mX_{2} = \text{DSConv}(\text{Conv}_3(\mX_{mid})) \\ 
        &\mX_3, \mX_4 = \text{Split}(\mX_{mid}) \\ 
        &\left. \begin{aligned}
        &\mX_5 = \text{ConvNeck}_1(\mX_4) + \mX_4 \\ 
        &\mX_6 = \text{ConvNeck}_2(\mX_5) + \mX_5 \\
        &\cdots \\
           &\mX_{4+n} = \text{ConvNeck}_n(\mX_{3+n}) + \mX_{3+n} \\
        \end{aligned} \right\} n
    \end{aligned}
    \right. ,
\end{equation}
where the channel number of $\mX_{mid}$ is $2c$. Whereas each of $ \mathbf{X}_{1}, \mathbf{X}_{2}, \ldots, \mathbf{X}_{4+n} $ features a channel count of $c$. 
Finally, we fuse and compress the semantic information of the three types of features through a concatenation operation followed by a $1 \times 1$ convolution to generate the $\mX_{out}$ with channel number $2c$, as follows:
\begin{equation}
    \mX_{out} = \text{Conv}_o(\mX_1 || \mX_2 || \cdots || \mX_{4+n}) .
\end{equation}

\subsection{Hypergraph-Based Cross-Level and Cross-Position Representation Network}
\begin{figure*}[!t]
    \centering
    \includegraphics[width=1.0\textwidth]{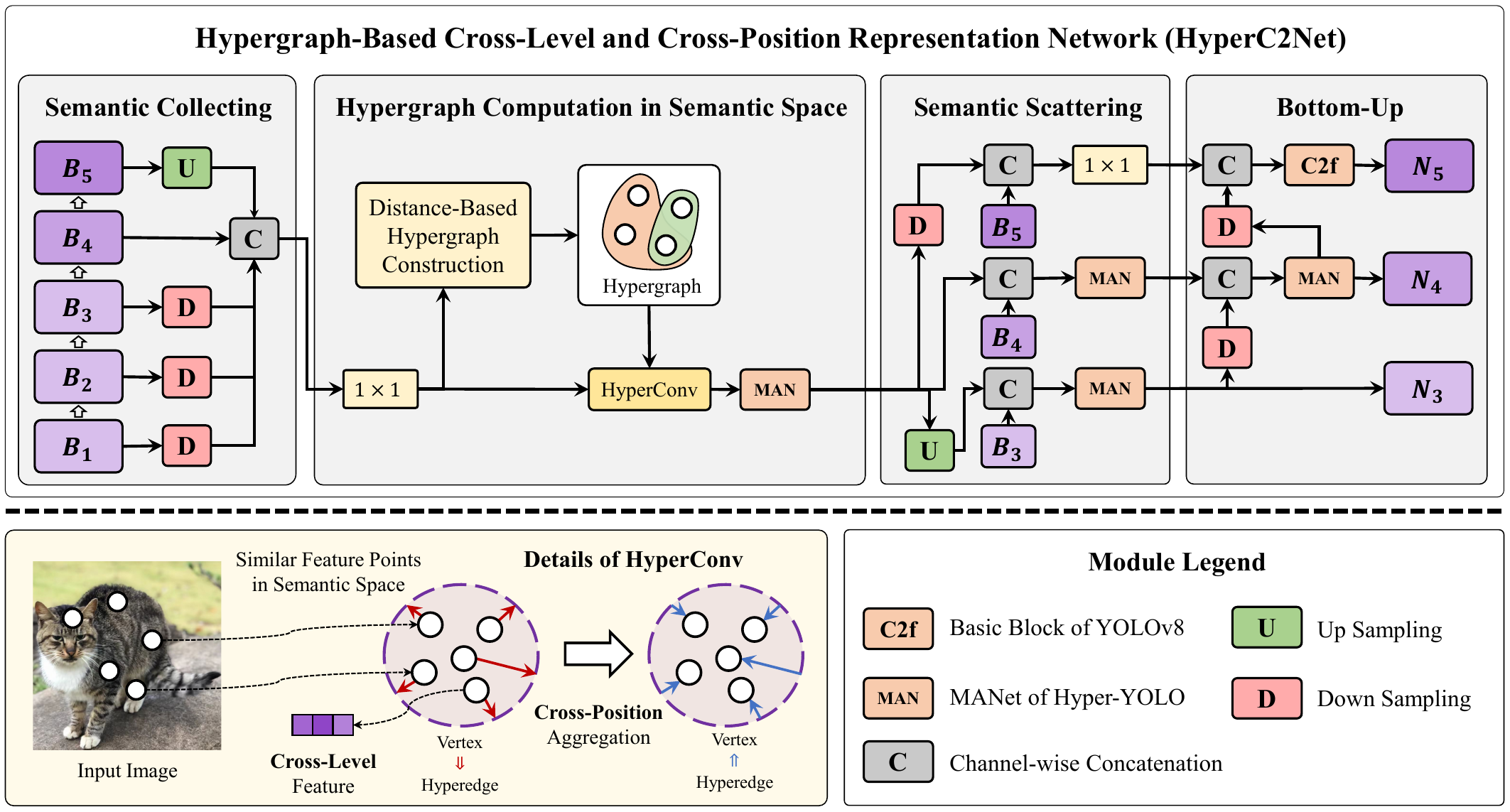}
    \caption{Illstration of the proposed Hypergraph-Based Cross-Level and Cross-Position Representation Network (HyperC2Net).}
    \label{fig:neck}
\end{figure*}

As for the neck of our Hyper-YOLO, in this subsection, to comprehensively fuse that cross-level and cross-position information from the backbone, we further propose the hypergraph-based cross-level and cross-position representation network (HyperC2Net), as shown in \cref{fig:neck}. HyperC2Net is an implementation of the proposed HGC-SCS framework, which is able to capture those potential high-order correlations in the semantic space.

\subsubsection{Hypergraph Construction.} 
As illustrated in \cref{app:fig:details}, our backbone is segmented into five discrete stages. The feature maps from these stages are represented as $\{\mB_1, \mB_2, \mB_3, \mB_4, \mB_5\}$. In an effort to harness hypergraph computation for elucidating the intricate high-order relationships among foundational features, we initiate the process by executing a channel-wise concatenation of the quintet of base features, thereby synthesizing cross-level visual features. 
A hypergraph $\gG = \{\gV, \gE\}$ is conventionally defined by its vertex set $ \gV $ and hyperedge set $ \gE $. In our approach, we deconstruct the grid-based visual features to constitute the vertex set $ \gV $ of a hypergraph. To model the neighborhood relationships within the semantic space, a distance threshold is used to construct an $\epsilon$-ball from each feature point, which will serve as a hyperedge, as illustrated \cref{fig:hg_con}. An $\epsilon$-ball is a hyperedge that encompasses all feature points within a certain distance threshold from a central feature point. The construction of the overall hyperedge set can be defined as $\gE = \{ ball(v, \epsilon) \mid v \in \gV \},$
where $ball(v, \epsilon) = \{u \mid ||\vx_u - \vx_v||_d < \epsilon, u \in \gV\}$ indicate the neighbor vertex set of the specified vertex $v$. $||\vx - \vy ||_d$ is the distance function.
In computations, a hypergraph $\gG$ is commonly represented by its incidence matrix $\mH$.


\subsubsection{Hypergraph Convolution.}
To facilitate high-order message passing on the hypergraph structure, we utilize a typical spatial-domain hypergraph convolution \cite{HGNNP} with extra residual connection to perform high-order learning on vertex features as follows:
\begin{equation}
\left\{
\begin{aligned}
    &\vx_e = \frac{1}{|\gN_v(e)|} \sum\nolimits_{v \in \gN_v(e)}\vx_v   {\bm \Theta} \\
    &\vx_v' = \vx_v + \frac{1}{|\gN_e(v)|} \sum\nolimits_{e \in \gN_e(v)} \vx_e
\end{aligned}
\right. ,
\end{equation}
where $\gN_v(e)$ and $\gN_e(v)$ are two neighbor indicate functions, as defined in \cite{HGNNP}: $\gN_v(e) = \{ v \mid v \in e, v \in \gV \}$ and $\gN_e(v) = \{ e \mid v \in e, e \in \gE \}$. ${\bm \Theta}$ is a trainable parameter. For computational convenience, the matrix formulation of the two-stage hypergraph message passing can be defined as:
\begin{equation}
    \text{HyperConv}(\mX, \mH) = \mX + \mD^{-1}_v \mH \mD^{-1}_e \mH^\top \mX {\bm \Theta},
\end{equation}
where $ \mD_v $ and $ \mD_e $ represent the diagonal degree matrices of the vertices and hyperedges, respectively.

\subsubsection{An Instance of HGC-SCS Framework.}
By combining the previously defined hypergraph construction and convolution strategies, we introduce a streamlined instantiation of the HGC-SCS framework, termed hypergraph-based cross-level and cross-position representation network (HyperC2Net), whose overarching definition is as follows:
\begin{equation}
\small
\nonumber
\left\{
\begin{aligned}
    &\mX_{mixed} = \mB_1 || \mB_2 || \mB_3 || \mB_4 || \mB_5 \\
    &\mX_{hyper} = \text{HyperConv}(\mX_{mixed}, \mH) \\
    &\mN_3, \mN_4, \mN_5 = \phi(\mX_{hyper}, \mB_3), \phi(\mX_{hyper}, \mB_4), \phi(\mX_{hyper}, \mB_4)
\end{aligned}
\right. ,
\end{equation}
where $\cdot||\cdot$ denotes the matrix concatenation operation. $\phi$ is the fusion function as illustrated in \cref{fig:neck} (semantic scattering module and bottom-up module). In our HyperC2Net, $ X_{mixed} $ intrinsically contains \textbf{cross-level} information, as it is a fusion of backbone features from multiple levels. Additionally, by deconstructing grid features into a set of feature points within the semantic space and constructing hyperedges based on distances, our approach permits high-order message passing among vertices at varying positions within the point set. This capability facilitates the capture of \textbf{cross-position} information, enriching the model's understanding of the semantic space.

\subsection{Comparison and Analysis}

Advancements in the YOLO series mainly concentrate on refinements to the backbone and neck components, with a specific focus on the backbone as a pivotal element of evolution with each successive YOLO iteration. For instance, the seminal YOLO\cite{yolo} framework introduced the DarkNet backbone, which has since undergone a series of enhancements, as exemplified by the ELAN (Efficient Layer Aggregation Network) module introduced in YOLOv7\cite{yolov7} and the C2f (Cross Stage Partial Connections with Feedback) module unveiled in YOLOv8\cite{yolov8}. These innovations have critically promoted the visual feature extraction prowess of the backbone architecture.

In contrast, our Hyper-YOLO model pivots the innovation axis towards the neck component's design. In the realm of neck architecture, leading-edge iterations such as YOLOv6\cite{yolov6}, YOLOv7\cite{yolov7}, and YOLOv8\cite{yolov8} have consistently incorporated the PANet\cite{PANet} (Path Aggregation Network) structure. Simultaneously, Gold-YOLO\cite{goldyolo} has adopted an inventive gather-distribute neck paradigm. In the following, we will compare HyperYOLO's HyperC2Net with these two classical neck architectures.

The PANet architecture, despite its efficacy in fusing multi-scale features via top-down and bottom-up pathways, remains constrained to the fusion of information across immediately contiguous layers. This adjacency-bound fusion modality inherently restricts the breadth of information integration within the network.
HyperC2Net, on the other hand, transcends this limitation by enabling direct fusion across the quintuple levels of features emanating from the backbone. This approach engenders a more robust and diversified information flow, curtailing the connectivity gap between features of varying depths. 
Notably, while the gather-distribute neck mechanism introduced by Gold-YOLO exhibits the capacity to assimilate information across multiple levels, it does not inherently account for cross-positional interactions within the feature maps.
The ingenuity of HyperC2Net lies in its utilization of hypergraph computations to capture the intricate high-order associations latent within feature maps. The hypergraph convolutions in the semantic domain facilitate a non-grid-constrained flow of information, empowering both cross-level and cross-positional high-order information propagation. Such a mechanism breaks the constrain of conventional grid structures, enabling a more nuanced and integrated feature representation.

The feature representations generated by HyperC2Net reflect a comprehensive consideration of both the semantic features provided by the original data backbone and the potential high-order structural features. Such an enriched feature representation is instrumental in achieving superior performance in object detection tasks. The ability of HyperC2Net to harness these intricate high-order relationships offers a significant advantage over conventional neck architectures like PANet and even recent innovations like the gather-distribute neck, underscoring the value of high-order feature processing in advancing the state-of-the-art in computer vision.

\section{Experiments}
\label{sec:exp}
\subsection{Experimental Setup}
\subsubsection{Datasets}
The Microsoft COCO dataset \cite{coco}, a benchmark for object detection, is employed to assess the efficacy of the proposed Hyper-YOLO model. In particular, the Train2017 subset is utilized for training purposes, while the Val2017 subset serves as the validation set. The performance evaluation of Hyper-YOLO is carried out on the Val2017 subset, with the results detailed in \cref{tab:main}.
\subsubsection{Compared Methods}
We select those advanced YOLO series methods, including YOLOv5 \cite{yolov5}, YOLOv6-3.0 \cite{yolov6}, YOLOv7 \cite{yolov7}, YOLOv8 \cite{yolov8}, Gold-YOLO \cite{goldyolo}, and YOLOv9 \cite{yolov9} for comparison. The default parameter configurations of their reported are adopted in our experiments.
\subsubsection{Our Hyper-YOLO Methods}
Our Hyper-YOLO is developed based on the four scales of YOLOv8 (-N, -S, -M, -L). Therefore, we modified the hyperparameters (number of convolutional layers, feature dimensions) for each stage of the Hyper-YOLO architecture, as shown in \cref{app:tab:neck}, resulting in Hyper-YOLO-N, Hyper-YOLO-S, Hyper-YOLO-M, and Hyper-YOLO-L. 
Considering that our Hyper-YOLO introduces high-order learning in the neck, which increases the number of parameters, we further reduced the parameters on the basis of Hyper-YOLO-N to form Hyper-YOLO-T. Specifically, in Hyper-YOLO-T's HyperC2Net, the last C2f in the Bottom-Up stage is replaced with a $1\times1$ convolution. Additionally, we noted that the latest YOLOv9 employs a new programmable gradient information transmission and prunes paths during inference to reduce parameters while maintaining accuracy. Based on YOLOv9, we developed Hyper-YOLOv1.1. Specifically, we replaced the neck of YOLOv9 with the HyperC2Net from Hyper-YOLO, thereby endowing YOLOv9 with the capability of high-order learning.
\subsubsection{Other Details}
To ensure an equitable comparison, we excluded the use of pre-training and self-distillation strategies for all methods under consideration, as outlined in \cite{yolov6} and \cite{goldyolo}. Furthermore, recognizing the potential influence of input image size on the evaluation, we standardized the input resolution across all experiments to $640 \times 640$ pixels, a common choice in the field. The evaluation is based on the standard COCO Average Precision (AP) metric. Additional implementation specifics are provided in \cref{app:sec:details} and \cref{app:sec:training}.

\begin{table*}[!t]
\centering
\caption{Comparison of state-of-the-art YOLO methods. 
The term ``\#Para.'' refers to the ``Number of Parameter'' within a model. Both Frames Per Second (FPS) and Latency were benchmarked under FP16 precision using a Tesla T4 GPU, consistent across all models with TensorRT 8.6.1.
During our practical speed evaluations, a notable observation was that TensorRT does not fully optimize the distance computation (\textit{torch.cdist()}), a crucial step in hypergraph construction. To maintain a fair comparison with other YOLO variants, we present additional results that isolate improvements to the backbone architecture alone, indicated by the symbol $\dag$. 
}
\label{tab:main}
\begin{tabular}{lcrrrrrrr}
\toprule
Method & Input Size & {$\text{AP}^{val}$} & $\text{AP}^{val}_{50}$ & \#Params. & FLOPs & FPS${\scriptstyle [bs=1]}$ & FPS$\scriptstyle [bs=32]$ & Latency$\scriptstyle [bs=1]$ \\ \midrule
YOLOv5-N \cite{yolov5} & 640 & 28.0\% & 45.7\% & 1.9 M & 4.5 G & 763 & 1158 & 1.3 ms \\
YOLOv5-S \cite{yolov5} & 640 & 37.4\% & 56.8\% & 7.2 M & 16.5 G & 455 & 606 & 2.2 ms \\
YOLOv5-M \cite{yolov5} & 640 & 45.4\% & 64.1\% & 21.2 M & 49.0 G & 220 & 267 & 4.6 ms \\
YOLOv5-L \cite{yolov5} & 640 & 49.0\% & 67.3\% & 46.5 M & 109.1 G & 133 & 148 & 7.5 ms \\ \midrule
YOLOv6-3.0-N \cite{yolov6} & 640 & 37.0\% & 52.7\% & 4.7 M & 11.4 G & 864 & 1514 & 1.2 ms \\
YOLOv6-3.0-S \cite{yolov6} & 640 & 44.3\% & 61.2\% & 18.5 M & 45.3 G & 380 & 581 & 2.6 ms \\
YOLOv6-3.0-M \cite{yolov6} & 640 & 49.1\% & 66.1\% & 34.9 M & 85.8 G & 198 & 263 & 5.1 ms \\
YOLOv6-3.0-L \cite{yolov6} & 640 & 51.8\% & 69.2\% & 59.6 M & 150.7 G & 116 & 146 & 8.6 ms \\ \midrule
Gold-YOLO-N \cite{goldyolo} & 640 & 39.6\% & 55.7\% & 5.6 M & 12.1 G & 694 & 1303 & 1.4 ms \\
Gold-YOLO-S \cite{goldyolo} & 640 & 45.4\% & 62.5\% & 21.5 M & 46.0 G & 331 & 530 & 3.0 ms \\
Gold-YOLO-M \cite{goldyolo} & 640 & 49.8\% & 67.0\% & 41.3 M & 87.5 G & 178 & 243 & 5.6 ms \\
Gold-YOLO-L \cite{goldyolo} & 640 & 51.8\% & 68.9\% & 75.1 M & 151.7 G & 107 & 139 & 9.3 ms \\ \midrule
YOLOv8-N \cite{yolov8} & 640 & 37.3\% & 52.6\% & 3.2 M & 8.7 G & 713 & 1094 & 1.4 ms \\
YOLOv8-S \cite{yolov8} & 640 & 44.9\% & 61.8\% & 11.2 M & 28.6 G & 395 & 564 & 2.5 ms \\
YOLOv8-M \cite{yolov8} & 640 & 50.2\% & 67.2\% & 25.9 M & 78.9 G & 181 & 206 & 5.5 ms \\
YOLOv8-L \cite{yolov8} & 640 & 52.9\% & 69.8\% & 43.7 M & 165.2 G & 115 & 127 & 8.7 ms \\ \midrule
YOLOv9-T\cite{yolov9} & 640 & 38.3\% & 53.1\% & 2.0 M & 7.7 G & 420 & 796 & 2.4 ms \\
YOLOv9-S\cite{yolov9} & 640 & 46.8\% & 63.4\% & 7.1 M & 26.4 G & 292 & 464 & 3.4 ms \\
YOLOv9-M\cite{yolov9} & 640 & 51.4\% & 68.1\% & 20.0 M & 76.3 G & 165 & 199 & 6.1 ms \\
YOLOv9-C\cite{yolov9} & 640 & 53.0\% & 70.2\% & 25.3 M & 102.1 G & 148 & 170 & 6.6 ms \\ \midrule
Hyper-YOLO-T & 640 & \textbf{38.5\%} & \textbf{54.5\%} & 3.1 M & 9.6 G & 404/692$^\dag$ & 644/1029$^\dag$ & 2.5/1.4$^\dag$ ms \\
Hyper-YOLO-N & 640 & \textbf{41.8\%} & \textbf{58.3\%} & 4.0 M & 11.4 G & 364/554$^\dag$ & 460/710$^\dag$ & 2.7/1.8$^\dag$ ms \\
Hyper-YOLO-S & 640 & \textbf{48.0\%} & \textbf{65.1\%} & 14.8 M & 39.0 G & 212/301$^\dag$ & 257/343$^\dag$ & 4.7/3.3$^\dag$ ms \\
Hyper-YOLO-M & 640 & \textbf{52.0\%} & \textbf{69.0\%} & 33.3 M & 103.3 G & 111/145$^\dag$ & 132/154$^\dag$ & 9.0/6.9$^\dag$ ms \\
Hyper-YOLO-L & 640 & \textbf{53.8\%} & \textbf{70.9\%} & 56.3 M & 211.0 G & 73/97$^\dag$ & 83/105$^\dag$ & 13.7/10.3$^\dag$ ms \\
Hyper-YOLOv1.1-T & 640 & \textbf{40.3\%} & \textbf{55.6\%} & 2.5 M & 10.8 G & 345 & 530 & 2.9 ms \\
Hyper-YOLOv1.1-S & 640 & \textbf{48.0\%} & \textbf{64.5\%} & 7.6 M & 29.9 G & 241 & 330 & 4.1 ms \\
Hyper-YOLOv1.1-M & 640 & \textbf{51.9\%} & \textbf{69.1\%} & 21.2 M & 87.4 G & 140 & 162 & 7.1 ms \\
Hyper-YOLOv1.1-C & 640 & \textbf{53.2\%} & \textbf{70.4\%} & 29.8 M & 115.5 G & 121 & 136 & 8.3 ms \\
\bottomrule
\end{tabular}
\end{table*}

\subsection{Results and Discussions}
The results of object detection on the COCO Val2017 validation set, as shown in \cref{tab:main}, lead to four main observations. 

Firstly, the proposed Hyper-YOLO method outperforms other models across all four scales. For instance, in terms of the $\text{AP}^{val}$ metric, Hyper-YOLO achieves a performance of 41.8\% at the -N scale, 48.0\% at the -S scale, 52.0\% at the -M scale, and 53.8\% at the -L scale. Compared to the Gold-YOLO, Hyper-YOLO shows an improvement of 2.2, 2.6, 2.2, and 2.0, respectively. When compared to YOLOv8, the improvements are 4.5, 3.1, 1.8, and 0.9, respectively. Compared to the YOLOv9, Hyper-YOLO shows an improvement of 3.5, 1.2, 0.6, and 0.8, respectively. These results validate the effectiveness of the Hyper-YOLO method.

Secondly, it is noteworthy that our method not only improves performance over Gold-YOLO but also reduces the number of parameters significantly. Specifically, there is a reduction of 28\% at the -N scale, 31\% at the -S scale, 19\% at the -M scale, and 25\% at the -L scale. The main reason for this is our HGC-SCS framework, which further introducs high-order learning in the semantic space comapred with the Gold-YOLO's gather-distribute mechanism. This allows our method to utilize the diverse information extracted by the backbone, including cross-level and cross-position information, more efficiently with fewer parameters.

Thirdly, considering that Hyper-YOLO shares a similar underlying architecture with YOLOv8, we found that the proposed Hyper-YOLO-T, compared to YOLOv8-N, achieved higher object detection performance (37.3 $\rightarrow$ 38.5 in terms of $AP^\text{val}$) with fewer parameters (3.2M $\rightarrow$ 3.1M). This demonstrates that the proposed HyperC2Net can achieve better feature representation learning through high-order learning, thereby enhancing detection performance.
Similarly, we compared Hyper-YOLOv1.1 with YOLOv9, as both use the same backbone architecture, with the only difference being that Hyper-YOLOv1.1 employs the hypergraph-based HyperC2Net as the neck. The results show that our Hyper-YOLOv1.1 demonstrated significant performance improvements: Hyper-YOLOv1.1-T outperformed YOLOv9-T by 2.0 $\text{AP}^{val}$, and Hyper-YOLOv1.1-S outperformed YOLOv9-S by 1.2 $\text{AP}^{val}$. This fair comparison using the same architecture at the same scale validates the effectiveness of the proposed high-order learning method in object detection tasks.

Finally, we observe that, compared to YOLOv8, the improvements brought by our Hyper-YOLO become more significant (from 0.9 to 4.5) as the model scale decreases (from -L to -N). This is because a smaller model scale weakens the feature extraction capability and the ability to obtain effective information from visual data. At this point, high-order learning becomes necessary to capture the latent high-order correlations in the semantic space of the feature map, enriching the features ultimately used for the detection head. Furthermore, high-order message propagation based on hypergraphs in the semantic space allows direct information flow between different positions and levels, enhancing the feature extraction capability of the base network with limited parameters.

\subsection{Ablation Studies on Backbone}
In this and the next subsection, taking into account the model's scale, we select the Hyper-YOLO-S to conduct ablation studies on the backbone and neck. 

\subsubsection{On Basic Block of Backbone.}
We conduct ablation experiments on the proposed MANet to verify the effectiveness of the mixed aggregation mechanism proposed in the basic block, as shown in \cref{tab:abl:block}. To ensure a fair comparison, we utilize the same PANet\cite{PANet} as the neck, used in YOLOv8\cite{yolov8}, so that the only difference between the two methods lies in the basic block. The experimental results clearly show that the proposed MANet outperforms the C2f module under the same neck across all metrics. This superior performance is attributed to the mixed aggregation mechanism, which integrates three classic structures, leading to a richer flow of information and thus demonstrating enhanced performance.

\begin{table}[!ht]
\caption{Ablation study on different basic blocks in the backbone.}
\label{tab:abl:block}
\centering
\begin{tabular}{lrrrrr}
\toprule
 & $\text{AP}^{val}$ & $\text{AP}^{val}_{50}$ & $\text{AP}^{s}$ & $\text{AP}^{m}$ & $\text{AP}^{l}$ \\
 & (\%) & (\%) & (\%) & (\%) & (\%) \\ \midrule
C2f(YOLOv8-S) & 44.9 & 61.7 & 25.9 & 49.7 & 61.0 \\
MANet(Ours) & \textbf{46.4} & \textbf{63.4} & \textbf{28.1} & \textbf{51.7} & \textbf{62.3} \\ \bottomrule
\end{tabular}
\end{table}

\subsubsection{On Kernel Size of Different Stages.}
We further conducted ablation experiments on the size of the convolutional kernels, an essential factor in determining the receptive field and the ability of a network to capture spatial hierarchies in data. In our experiments, $ k_i $ represents the kernel size of the MANet used at the $ i $-th stage. Since our MANet begins to utilize mixed aggregation starting from the second stage, the configuration of $ k $ in our experiments is denoted as $ [k_2, k_3, k_4, k_5] $. Experimental results are presented in \cref{tab:abl:kernel}.
The experimental results indicate that increasing the size of the convolutional kernels from 3 to 5 can indeed enhance the model's accuracy. However, for small-scale and medium-scale object detection, the accuracy does not necessarily improve compared to a mixture of different kernel sizes, and it also results in a larger number of parameters. 
Therefore, taking into account a balance between performance and the number of parameters, our Hyper-YOLO ultimately selects the $[3, 5, 5, 3]$ configuration as the optimal setting for the convolutional kernel sizes in our MANet.

\begin{table}[!ht]
\centering
\caption{Ablation study on different kernel size settings.}
\label{tab:abl:kernel}
\begin{tabular}{lrrrrr}
\toprule
\multirow{2}{*}{$[k_2, k_3, k_4, k_5]$} & $\text{AP}^{val}$ & $\text{AP}^{val}_{50}$ & $\text{AP}^{s}$ & $\text{AP}^{m}$ & $\text{AP}^{l}$ \\ 
& (\%) & (\%) & (\%) & (\%) & (\%) \\ \midrule 
$[3, 3, 3, 3]$ & 46.3 & 63.3 & 27.2 & 51.1 & 62.6 \\ 
$[5, 5, 5, 5]$ & \textbf{46.6} & \textbf{63.5} & 27.5 & 51.6 & \textbf{63.1} \\ 
$[3, 5, 5, 3]$ & 46.4 & 63.4 & \textbf{28.1} & \textbf{51.7} & 62.3 \\ \bottomrule
\end{tabular}
\end{table}

\subsection{Ablation Studies on Neck}
\subsubsection{High-Order \textit{vs.} Low-Order Learning in HGC-SCS Framework}
The core of the HGC-SCS framework lies in the semantic space's Hypergraph Computation, which allows for high-order information propagation among feature point sets. We conduct ablation studies to evaluate its effectiveness by simplifying the hypergraph into a graph for low-order learning, as shown in \cref{tab:abl:learning}. In this case, the graph is constructed by connecting the central node with its neighbors within an $\epsilon$-ball. The graph convolution operation used \cite{gcn} is the classic one: $\hat{\mA} = \mD_v^{-1/2}\mA \mD_v^{-1/2} + \mI$, where $\mD_v$ is the diagonal degree matrix of the graph adjacency matrix $\mA$.
Additionally, we include a configuration with no correlation learning at all: ``None''. The experimental results, as presented in \cref{tab:abl:learning}, reveal that high-order learning demonstrates superior performance compared to the other two methods. Theoretically, low-order learning can be considered a subset \cite{HGM2R} of high-order learning but lacks the capability to model complex correlation. High-order learning, on the other hand, possesses a more robust correlation modeling capability, which corresponds with a higher performance ceiling. As a result, it tends to achieve better performance more easily.

\begin{table}[!ht]
\centering
\caption{Ablation study on different enhancement strategies.}
\label{tab:abl:learning}
\begin{tabular}{lrrrrr}
\toprule
\multirow{2}{*}{Hypergraph Computation} & $\text{AP}^{val}$ & $\text{AP}^{val}_{50}$ & $\text{AP}^{s}$ & $\text{AP}^{m}$ & $\text{AP}^{l}$ \\
& (\%) & (\%) & (\%) & (\%) & (\%) \\ \midrule
None & 46.4 & 63.4 & 28.1 & 51.7 & 62.3 \\
Low-Order Learning & 47.6 & 64.8 & 29.1 & 53.1 & 63.7 \\
High-Order Learning & \textbf{48.0} & \textbf{65.1} & \textbf{29.9} & \textbf{53.2} & \textbf{64.6} \\ \bottomrule
\end{tabular}
\end{table}

\subsubsection{On the Semantic Collecting Phase}
The first phase of the HGC-SCS framework is Semantic Collecting, which determines the total amount of information fed into the semantic space for hypergraph computation. We performed ablation studies on this phase, as shown in \cref{tab:abl:collecting}, using three different configurations that select 3, 4, or 5 levels of feature maps for input.
The experimental results reveal that a greater number of feature maps can bring more abundant semantic space information. This enhanced information richness allows the hypergraph to fully exploit its capability in modeling complex correlation. Consequently, the input configuration with 5 feature maps achieved the best performance.
This outcome suggests that the model can benefit from a more comprehensive representation of the input data when more levels of feature maps are integrated. The inclusion of more feature maps likely introduces a broader range of semantic meaning and details from the visual input, enabling the hypergraph to establish higher-order connections that reflect a more complete understanding of the scene. Therefore, the configuration that incorporates 5 feature maps is preferred for maximizing the potential of hypergraph-based complex correlation modeling.

\begin{table}[!ht]
\centering
\caption{Ablation study on the number of input levels.
}
\label{tab:abl:collecting}
\begin{tabular}{lrrrrr}
\toprule
\multirow{2}{*}{Semantic Collecting Set} & $\text{AP}^{val}$ & $\text{AP}^{val}_{50}$ (\%) & $\text{AP}^{s}$ & $\text{AP}^{m}$ & $\text{AP}^{l}$ \\
& (\%) & (\%) & (\%) & (\%) & (\%) \\ \midrule
$\{\mB_3, \mB_4, \mB_5\}$ & 47.5 & 64.6 & 28.9 & 52.6 & 63.8 \\
$\{\mB_2, \mB_3, \mB_4, \mB_5\}$ & 47.8 & 65.0 & 28.4 & 53.1 & 64.2 \\
$\{\mB_1, \mB_2, \mB_3, \mB_4, \mB_5\}$ & \textbf{48.0} & \textbf{65.1} & \textbf{29.9} & \textbf{53.2} & \textbf{64.6} \\ \bottomrule
\end{tabular}
\end{table}

\subsubsection{On Hypergraph Construction of Hypergraph Computation Phase}
Further ablation experiments are conducted to examine the effect of the distance threshold used in the construction of the hypergraph, with the results shown in \cref{tab:abl:threshold}. Compared to the configuration ``None'' where hypergraph computation is not introduced, the introduction of hypergraph computation leads to a significant overall performance improvement. It is also observed that the performance of the target detection network is relatively stable across a range of threshold values from 7 to 9, with only minor variations.
However, there is a performance decline at the thresholds of 6 and 10. This decline can be attributed to the number of connected nodes directly affecting the smoothness of features in the semantic space. A higher threshold may lead to a more connected hypergraph, where nodes are more likely to share information, potentially leading to over-smoothing of the features. Conversely, a lower threshold may result in a less connected hypergraph that cannot fully exploit the high-order relationships among features.
Therefore, our Hyper-YOLO uses the distance threshold 8 for construction. The precise value would be determined based on empirical results, balancing the need for a richly connected hypergraph against the risk of over-smoothing or under-connecting the feature representation.

\begin{table}[!ht]
\centering
\caption{Ablation study on the threshold of hypergraph construction.}
\label{tab:abl:threshold}
\begin{tabular}{crrrrr}
\toprule
\multirow{2}{*}{Distance Threshold} & $\text{AP}^{val}$ & $\text{AP}^{val}_{50}$ & $\text{AP}^{s}$ & $\text{AP}^{m}$ & $\text{AP}^{l}$ \\
 & (\%) & (\%) & (\%) & (\%) & (\%) \\ \midrule
None & 46.3 & 63.5 & 26.9 & 51.6 & 62.6 \\ 
6 & 47.6 & 64.6 & 28.6 & 52.7 & 64.2 \\
7 & 47.8 & 65.0 & 29.4 & 53.3 & 64.0 \\
\textbf{8} & \textbf{48.0} & \textbf{65.1} & \textbf{29.9} & 53.2 & \textbf{64.6} \\
9 & 47.8 & 64.9 & 29.2 & \textbf{53.4} & 64.5 \\
10 & 47.7 & 65.1 & 28.2 & 53.0 & 63.7 \\ 
\bottomrule
\end{tabular}
\end{table}

\subsection{More Ablation Studies}
In this subsection, we conduct thorough ablation studies to assess the impact of backbone and neck enhancements in Hyper-YOLO across four different model scales, with detailed results presented in \cref{app:abl:more}. The baseline performance of YOLOv8 is placed at the top of the table. The middle part of the table introduces our HyperYOLO models that incorporate only the backbone enhancement. At the bottom, we feature the fully augmented HyperYOLO models, which benefit from both backbone and neck enhancements. Based on experimental results in \cref{app:abl:more}, we have three observations.

\begin{table*}
\caption{Ablation studies on different scale models.}
\label{app:abl:more}
\centering
\begin{tabular}{lrrrrrrrrrrrr}
\toprule
{Method} & {$\text{AP}^{val}$} & {$\text{AP}^{val}_{50}$} & {$\text{AP}^s$} & {$\text{AP}^m$} & {$\text{AP}^l$} & {\#Params.} & {FLOPs} & FPS ${\scriptstyle [bs=1]}$ & FPS $\scriptstyle [bs=32]$ & Latency $\scriptstyle [bs=1]$ \\
 \midrule
YOLOv8-N & 37.3\% & 52.3\% & 18.7\% & 40.9\% & 53.3\% & 3.2 M & 8.7 G & 713 & 1094 & 1.4 ms \\
YOLOv8-S & 44.9\% & 61.7\% & 25.9\% & 49.7\% & 61.0\% & 11.2 M & 28.6 G & 395 & 564 & 2.5 ms \\
YOLOv8-M & 50.2\% & 67.1\% & 32.3\% & 55.6\% & 66.5\% & 25.9 M & 78.9 G & 181 & 206 & 5.5 ms \\
YOLOv8-L & 52.9\% & 69.6\% & 35.1\% & 57.9\% & 69.8\% & 43.7 M & 165.2 G & 115 & 127 & 8.7 ms \\ \midrule
\multicolumn{11}{c}{Backbone Enhancement} \\ \midrule
HyperYOLO-N & 39.9\% & 56.2\% & 20.8\% & 44.3\% & 55.5\% & 3.5 M & 9.8 G & 554 & 710 & 1.8 ms \\
HyperYOLO-S & 46.4\% & 63.4\% & 28.1\% & 51.7\% & 62.3\% & 12.7 M & 32.6 G & 301 & 343 & 3.3 ms \\
HyperYOLO-M & 51.0\% & 67.9\% & 32.7\% & 56.8\% & 67.9\% & 28.2 M & 86.8 G & 145 & 154 & 6.9 ms \\
HyperYOLO-L & 53.0\% & 70.0\% & 35.7\% & 58.8\% & 69.5\% & 46.4 M & 177.8 G & 97 & 105 & 10.3 ms \\ \midrule
\multicolumn{11}{c}{Backbone\&Neck Enhancement} \\ \midrule
HyperYOLO-N & 41.8\% & 58.3\% & 22.2\% & 46.4\% & 58.7\% & 4.0 M & 11.4 G & 364 & 460 & 2.7 ms \\
HyperYOLO-S & 48.0\% & 65.1\% & 29.9\% & 53.2\% & 64.6\% & 14.8 M & 39.0 G & 212 & 257 & 4.7 ms \\
HyperYOLO-M & 52.0\% & 69.0\% & 34.6\% & 57.9\% & 68.7\% & 33.3 M & 103.3 G & 111 & 132 & 9.0 ms \\
HyperYOLO-L & \textbf{53.8\%} & \textbf{70.9\%} & \textbf{36.7\%} & \textbf{59.9\%} & \textbf{70.2\%} & 56.3 M & 211.0 G & 73 & 83 & 13.7 ms \\ \bottomrule
\end{tabular}
\end{table*}

Firstly, the adoption of both individual and combined enhancements significantly boosts performance for the -N, -S, and -M models, validating the effectiveness of our proposed modifications. Secondly, the impact of each enhancement appears to be scale-dependent. As we progress from -N to -S, -M, and -L models, the incremental performance gains due to the backbone improvement gradually decrease from 2.6 to 1.5, 0.8, and finally 0.1. In contrast, the neck enhancement consistently contributes more substantial improvements across these scales, with respective gains of 1.9, 1.6, 1.0, and 0.8. This suggests that while the benefits of an expanded receptive field and width scaling in the backbone are more pronounced in smaller models, the advanced HyperC2Net neck provides a more uniform enhancement by enriching the semantic content and boosting object detection performance across the board.
Thirdly, when focusing on small object detection ($\text{AP}^s$), the HyperYOLO-L model with both backbone and neck enhancements achieves a notable increase of 1.6, whereas just the backbone enhancement leads to a 0.6 improvement. This underscores the potential of hypergraph modeling, particularly within the neck enhancement, to capture the complex relationships among small objects and significantly improve detection in these challenging scenarios.

\subsection{More Evaluation on Instance Segmentation Task}
\begin{table}
\caption{Experimental results on instance segmentation task.}
\label{tab:seg}
\centering
\begin{tabular}{lrrrr}
\toprule
\multirow{2}{*}{Methods} & $\text{AP}^{box}$ & $\text{AP}^{mask}$ & Params. & FLOPs \\
 & (\%) & (\%) & (M) & (G) \\ \midrule
YOLOv8-N-seg & 36.7 & 30.5 & 3.4 & 12.6 \\
YOLOv8-S-seg & 44.6 & 36.8 & 11.8 & 42.6 \\
YOLOv8-M-seg & 49.9 & 40.8 & 27.3 & 110.2 \\
YOLOv8-L-seg & 52.3 & 42.6 & 46.0 & 220.5 \\ \midrule
HyperYOLO-N-seg & \textbf{41.4} & \textbf{33.8} & 4.3 & 15.3 \\
HyperYOLO-S-seg & \textbf{47.9} & \textbf{39.1} & 15.5 & 53.0 \\
HyperYOLO-M-seg & \textbf{52.1} & \textbf{42.1} & 34.7 & 134.6 \\ 
HyperYOLO-L-seg & \textbf{53.7} & \textbf{43.3} & 58.6 & 266.3 \\ \bottomrule
\end{tabular}
\end{table}

We extend the application of Hyper-YOLO to the instance segmentation task on the COCO dataset, ensuring a direct comparison with its predecessor, YOLOv8, by adopting a consistent approach in network modification: replacing the detection head with a segmentation head. Experimental results are shown in \cref{tab:seg}.

The empirical results clearly illustrate that Hyper-YOLO attains remarkable performance enhancements. For $\text{AP}^{box}$, Hyper-YOLO shows an impressive increase of 4.7 AP for the -N variant, 3.3 AP for the -S variant, 2.2 AP for the -M variant, and 1.4 AP for the -L variant. Similarly, for $\text{AP}^{mask}$, Hyper-YOLO exhibits significant improvements, with gains of 3.3 AP for -N, 2.3 AP for -S, 1.3 AP for -M, and 0.7 AP for -L. These results underscore the effectiveness of the advancements integrated into Hyper-YOLO.

\subsection{Visualization of High-Order Learning in Object Detection}
\begin{figure}
    \centering
    \includegraphics[width=\linewidth]{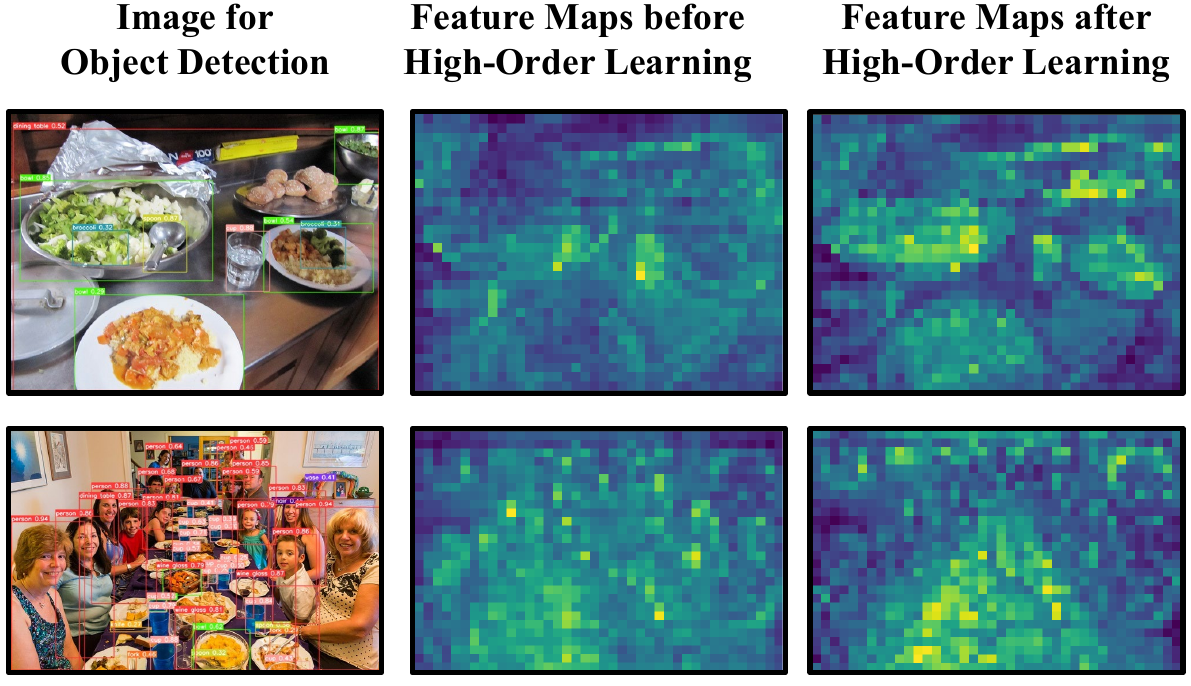}
    \caption{Visualization of feature maps before and after high-order learning.}
    \label{fig:vis_ft}
\end{figure}

In our paper, we have provided a mathematical rationale explaining how the hypergraph-based neck can transcend the limitations of traditional neck designs, which typically rely on grid-like neighborhood structures for message propagation within feature maps. This design enables advanced high-order message propagation across the semantic spaces of the features.
To further substantiate the effectiveness of our hypergraph-based neck, we have included visualizations in the revised manuscript, as shown in \cref{fig:vis_ft}. These visualizations compare feature maps before and after applying our HyperConv layer. It is evident from these images that there is a consistent reduction in attention to semantically similar backgrounds, such as skies and grounds, while maintaining focus on foreground objects across various scenes. This demonstrates that HyperConv, through hypergraph computations, aids the neck in better recognizing semantically similar objects within an image, thus supporting the detection head in making more consistent decisions.

\section{Conclusion}
\label{sec:con}
In this paper, we presented Hyper-YOLO, a groundbreaking object detection model that integrates hypergraph computations with the YOLO architecture to harness the potential of high-order correlations in visual data. By addressing the inherent limitations of traditional YOLO models, particularly in the neck design's inability to effectively integrate features across different levels and exploit high-order relationships, we have significantly advanced the SOTA in object detection. 
Our contributions set a new benchmark for future research and development in object detection frameworks and pave the way for further exploration into the integration of hypergraph computations within visual architectures based on our HGC-CSC framework.



\bibliographystyle{IEEEtran}
\bibliography{IEEEabrv,main}

\begin{IEEEbiography}[{
\includegraphics[width=1in,height=1.25in,clip,keepaspectratio]{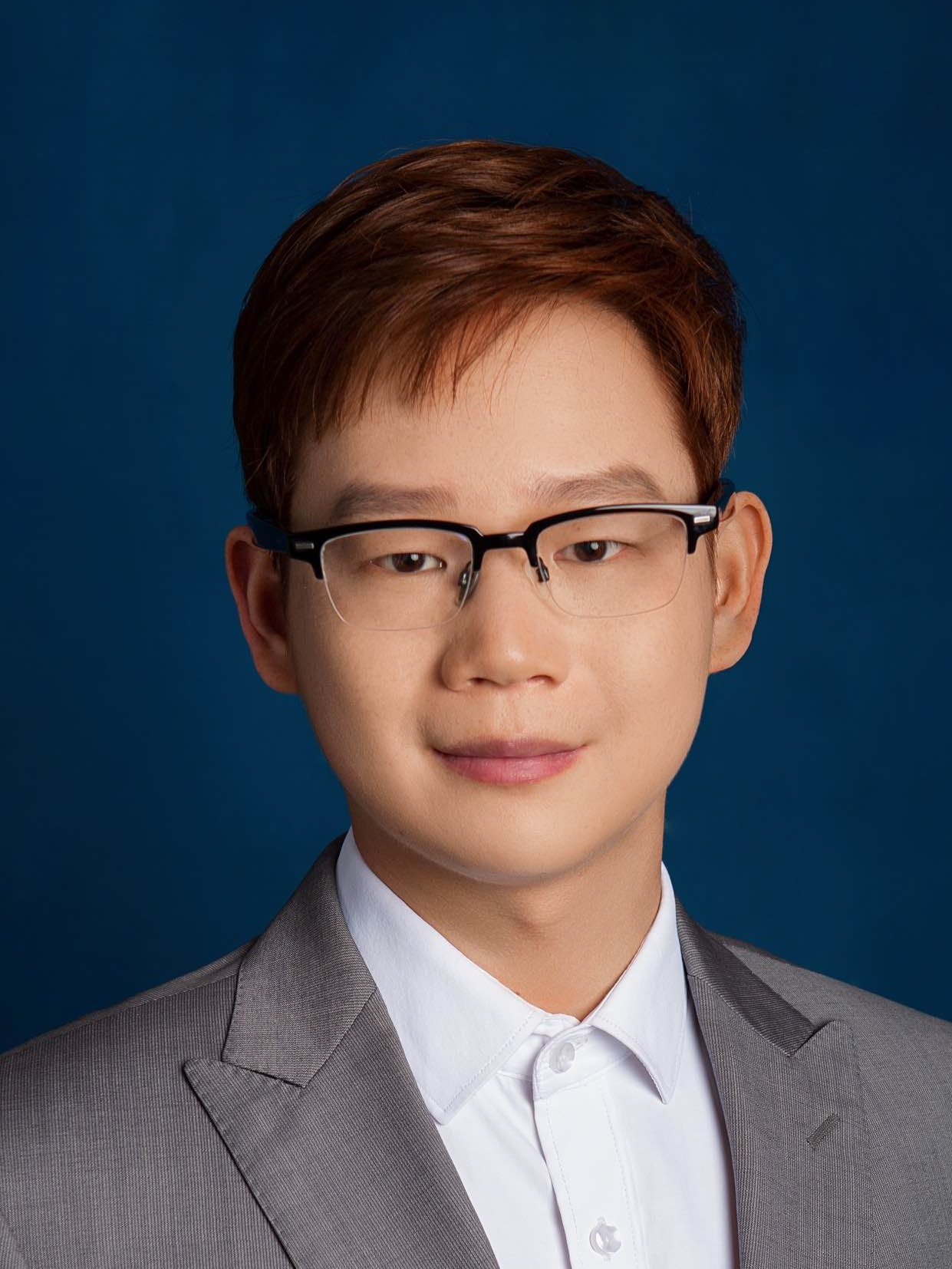}}]{Yifan Feng} 
received the BE degree in computer science and technology from Xidian University, Xi'an, China, in 2018, and the MS degree from Xiamen University, Xiamen, China, in 2021. He is currently working toward the PhD degree from the School of Software, Tsinghua University, Beijing, China. His research interests include hypergraph neural networks, machine learning, and pattern recognition.
\end{IEEEbiography}
\begin{IEEEbiography}[{
\includegraphics[width=1in,height=1.25in,clip,keepaspectratio]{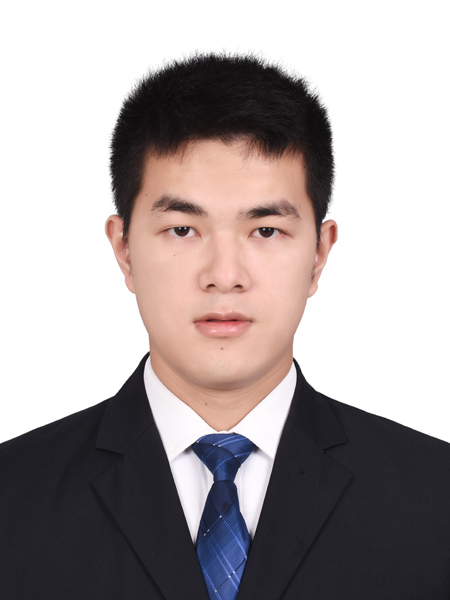}}]{Jiangang Huang} 
received the BE degree in software engineering from Xi’an Jiaotong University, Xi’an, China, in 2022. He is currently working toward the master’s degree in the same field from Xi’an Jiaotong University. His research interests include object detection, software engineering, and artificial intelligence.
\end{IEEEbiography}
\begin{IEEEbiography}[{
\includegraphics[width=1in,height=1.25in,clip,keepaspectratio]{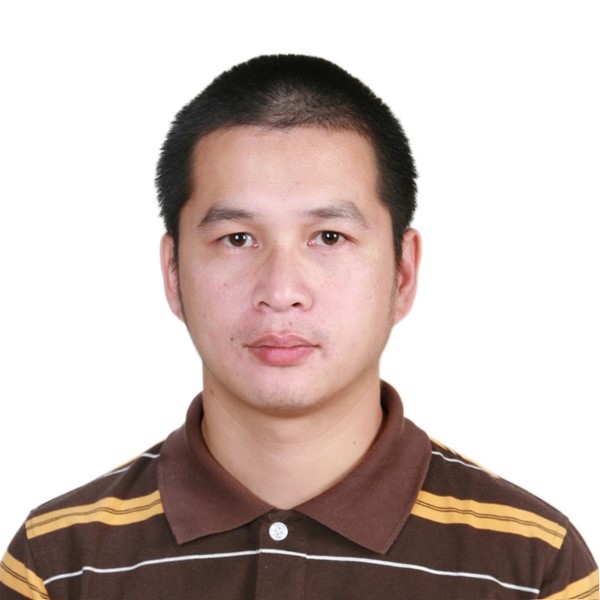}}]{Shaoyi Du} 
received double Bachelor degrees in computational mathematics and in computer science in 2002 and received his M.S. degree in applied mathematics in 2005 and Ph.D. degree in pattern recognition and intelligence system from Xi’an Jiaotong University, China in 2009. He is a professor at Xi’an Jiaotong University. His research interests include computer vision, machine learning and pattern recognition.
\end{IEEEbiography}
\vfill
\begin{IEEEbiography}[{\vspace{-0.8cm}
\includegraphics[width=1in,height=1.25in,clip,keepaspectratio]{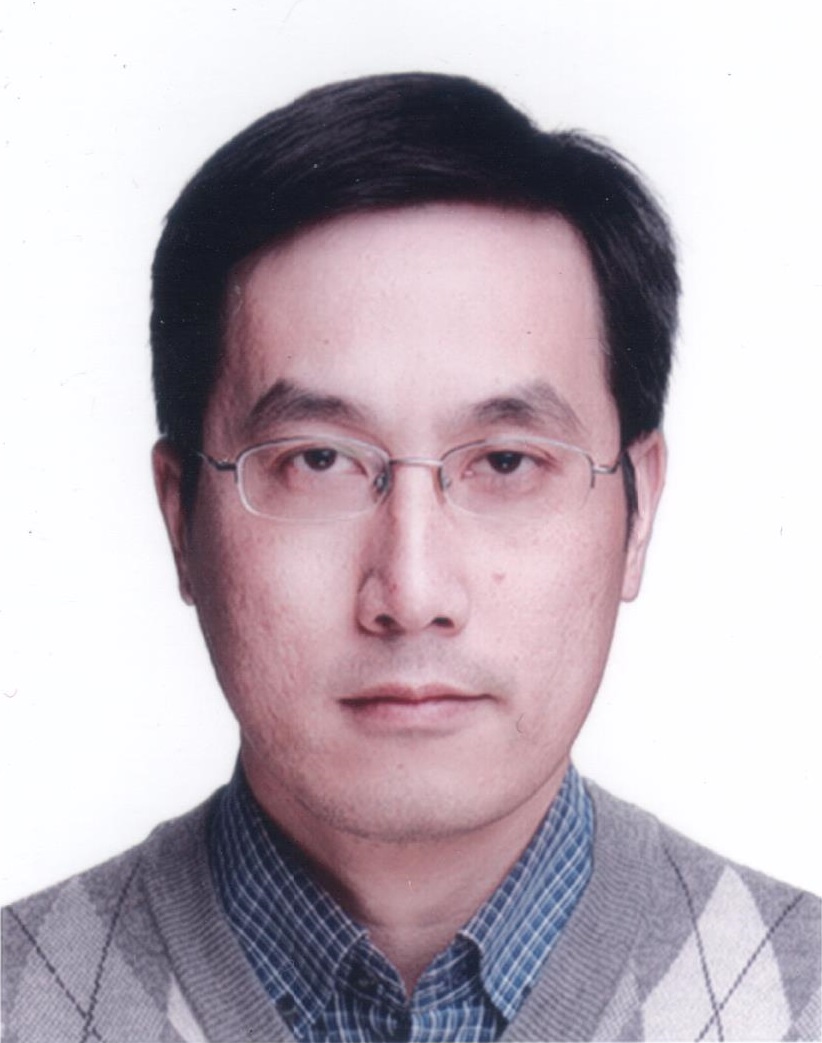}}]{Shihui Ying (M'11)} 
received the B.Eng. degree in mechanical engineering and the Ph.D. degree in applied mathematics from Xi'an Jiaotong University, Xi'an, China, in 2001 and 2008, respectively. He is currently a Professor with the Department of Mathematics, School of Science, Shanghai University, Shanghai, China. His current research interests include geometric theory and methods for machine intelligence and medical image analysis.
\end{IEEEbiography}
\begin{IEEEbiography}[{
\includegraphics[width=1in,height=1.25in,clip,keepaspectratio]{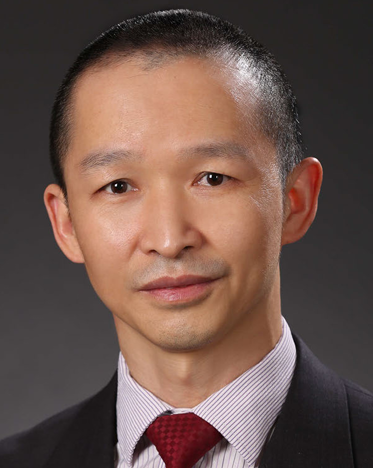}}]{Jun-Hai Yong} 
received the B.S. and Ph.D. degrees in computer science from Tsinghua University, Beijing, China, in 1996 and 2001, respectively. He held a visiting researcher position with the Department of Computer Science, Hong Kong University of Science and Technology in 2000. He was a Post-Doctoral Fellow with the Department of Computer Science, University of Kentucky, from 2000 to 2002. He is currently a Professor with the School of Software, Tsinghua University. His main research interests include computer-aided design and computer graphics. He received a lot of awards, such as the National Excellent Doctoral Dissertation Award, the National Science Fund for Distinguished Young Scholars, the Best Paper Award of the ACM SIGGRAPH Eurographics Symposium on Computer Animation, the Outstanding Service Award as an Associate Editor of the Computers and Graphics journal by Elsevier, and several National Excellent Textbook Awards.
\end{IEEEbiography}
\begin{IEEEbiography}[{
\includegraphics[width=1in,height=1.25in,clip,keepaspectratio]{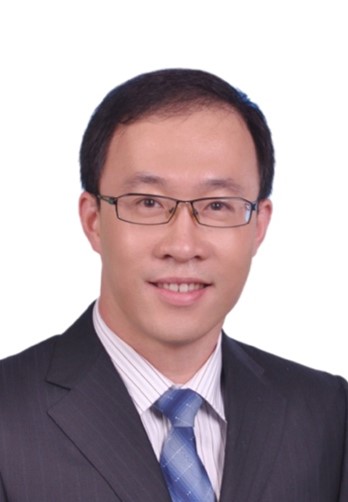}}]{Yipeng Li} 
received the B.S. and M.S. degrees in electronic engineering from the Harbin Institute of Technology, Harbin, China, and the Ph.D. degree in electronic engineering from Tsinghua University, Beijing, China, in 2003, 2005, and 2011, respectively. He is currently an Assistant Researcher with the Department of Automation, Tsinghua University. His current research interests include UAV vision-based autonomous navigation, 3-D reconstruction of natural environment, complex systems theory, and Internet applications analysis.
\end{IEEEbiography}
\begin{IEEEbiography}[{
\includegraphics[width=1in,height=1.25in,clip,keepaspectratio]{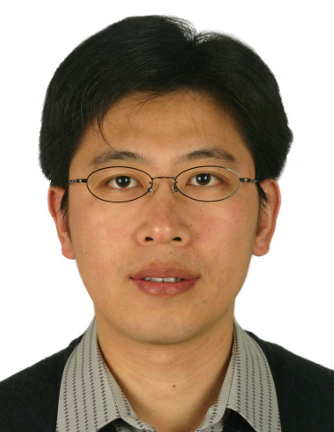}}]{Guiguang Ding} 
is currently a Distinguished Researcher with the School of Software, Tsinghua University; a Ph.D. Supervisor; an Associate Dean of the School of Software, Tsinghua University; and the Deputy Director of the National Research Center for Information Science and Technology. His research interests mainly focus on visual perception, theory and method of efficient retrieval and weak supervised learning, neural network compression of vision task under edge computing and power lim ited scenes, visual computing systems, and platform developing. He was the Winner of the National Science Fund for Distinguished Young Scholars.
\end{IEEEbiography}
\vfill
\begin{IEEEbiography}[{
\includegraphics[width=1in,height=1.25in,clip,keepaspectratio]{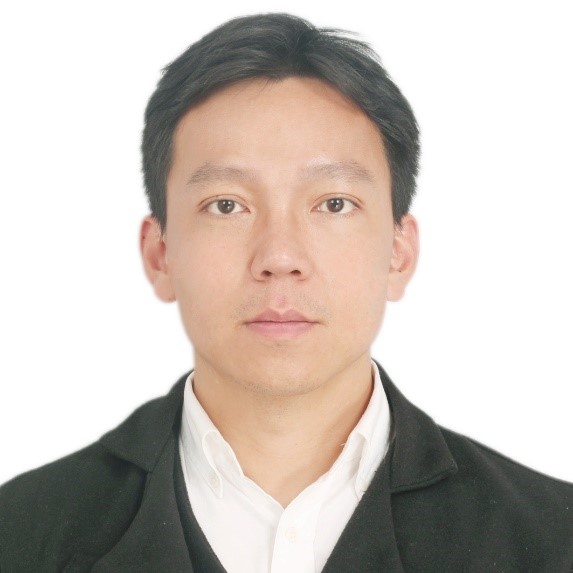}}]{Rongrong Ji} 
is currently a Professor and the Director of the Intelligent Multimedia Technology Laboratory, School of Informatics, Xiamen University, Xiamen, China. His work mainly focuses on innovative technologies for multimedia signal processing, computer vision, and pattern recognition, with over 100 papers published in international journals and conferences. He serves as an Associate/Guest Editor for international journals and magazines, such as Neurocomputing, Signal Processing, and Multimedia Systems.
\end{IEEEbiography}
\begin{IEEEbiography}[{\vspace{-1cm} 
\includegraphics[width=1in,height=1.25in,clip,keepaspectratio]{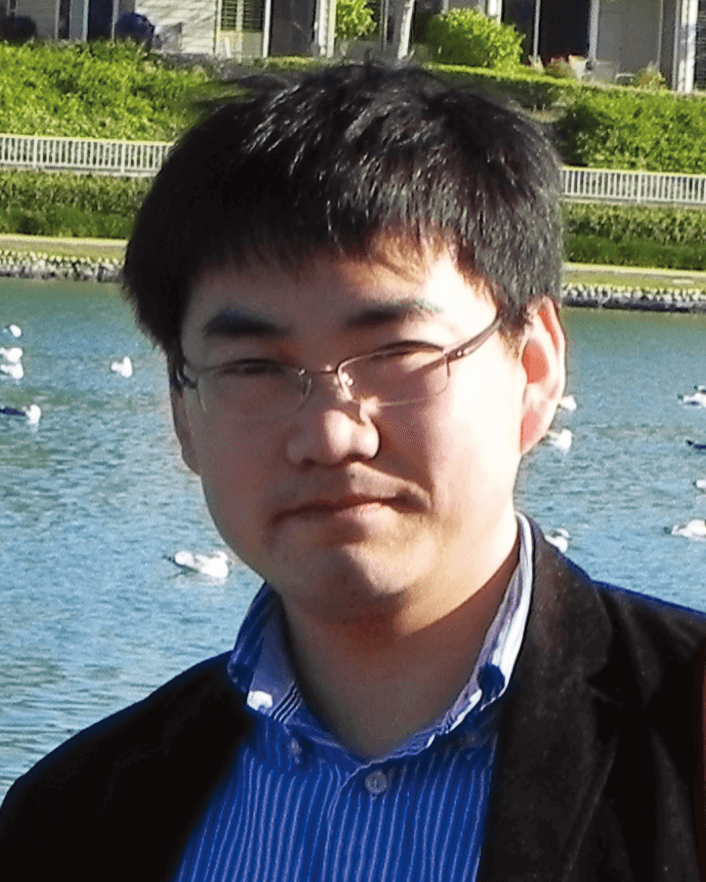}}]{Yue Gao}
is an associate professor with the School of Software, Tsinghua University. He received the B.S. degree from the Harbin Institute of Technology, Harbin, China, and the M.E. and Ph.D. degrees from Tsinghua University, Beijing, China.
\end{IEEEbiography}
\vfill

\clearpage

\setcounter{table}{0}  
\setcounter{figure}{0}
\renewcommand{\thetable}{S\arabic{table}}
\renewcommand{\thefigure}{S\arabic{figure}}

\appendices

\begin{figure*}[ht]
    \centering
    \includegraphics[width=\linewidth]{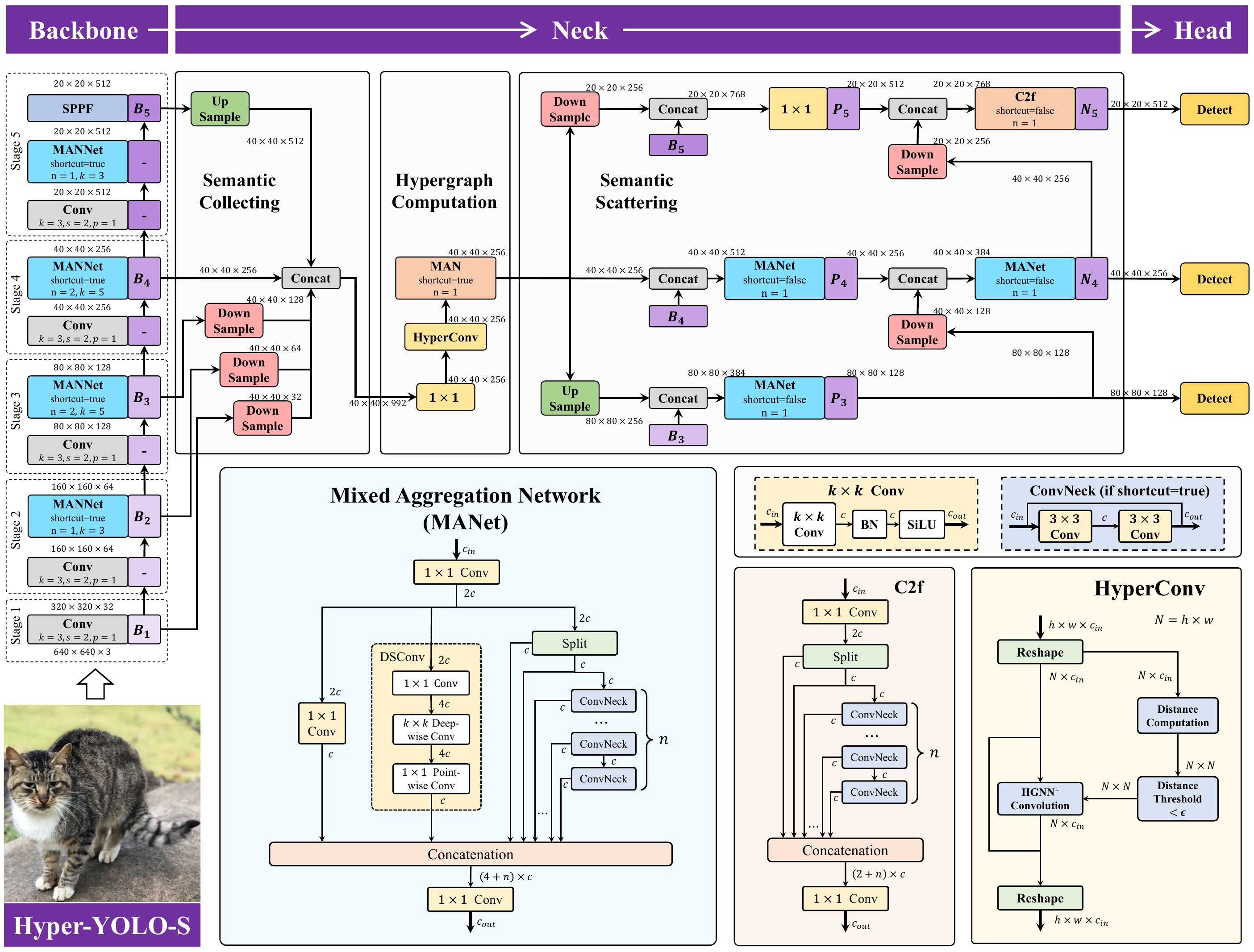}
    \caption{The detailed configuration of our Hyper-YOLO-S.}
    \label{app:fig:details}
\end{figure*}

\section{Implemental Details of Hyper-YOLO}
\label{app:sec:details}
In this section, we detail the implementation of our proposed models: Hyper-YOLO-N, Hyper-YOLO-S, Hyper-YOLO-M, and Hyper-YOLO-L. These models are developed upon the PyTorch\footnote{\url{https://pytorch.org/}}. In line with the configuration established by YOLOv8 \cite{yolov8}, our models share analogous architectures and loss functions, with the notable exception of incorporating MANet and HyperC2Net. An efficient decoupled head has been integrated for precise object detection. The specific configurations of the Hyper-YOLO-S are depicted in \cref{app:fig:details}.

\begin{table}[]
\caption{Configuration of Hyper-YOLO's backbone. The Stage $1-5$ refer to \cref{app:fig:details}.}
\centering
\label{app:tab:backbone}
\begin{tabular}{lccccccccccccccccc}
\toprule
& \multicolumn{10}{c}{MANet} \\ 
& \multicolumn{2}{c}{Stage 1} & \multicolumn{2}{c}{Stage 2} & \multicolumn{2}{c}{Stage 3} & \multicolumn{2}{c}{Stage 4} & \multicolumn{2}{c}{Stage 5} \\ 
& $n$ & $k$ & $n$ & $k$ & $n$ & $k$ & $n$ & $k$ & $n$ & $k$ \\ \midrule
Hyper-YOLO-N & - & - & 1 & 3 & 2 & 5 & 2 & 5 & 1 & 3 \\
Hyper-YOLO-S & - & - & 1 & 3 & 2 & 5 & 2 & 5 & 1 & 3 \\
Hyper-YOLO-M & - & - & 2 & 3 & 4 & 5 & 4 & 5 & 2 & 3 \\
Hyper-YOLO-L & - & - & 3 & 3 & 6 & 5 & 6 & 5 & 3 & 3 \\ \bottomrule
\end{tabular}
\end{table}

\subsection{Backbone} 
The backbone of HyperYOLO, detailed in \cref{app:tab:backbone}, has been updated from its predecessor, with the C2f module being replaced by the MANet module, maintaining the same number of layers as in YOLOv8 \cite{yolov8}, structured as $[3,6,6,3]$. The channel counts for each stage are kept consistent with those in YOLOv8, with the only change being the module swap. The MANet employs depthwise separable convolutions with an increased channel count, where a $2c$ input is expanded to a $4c$ output (with $2c$ equivalent to $c_{out}$).

In addition to these adjustments, the hyperparameters $k$ and $n$ for the four stages are set to $[3,5,5,3]$ and $[3,6,6,6]$ $\times$ depth, respectively. The depth multiplier varies across the different scales of the model, being set to $1/3$, $1/3$, $2/3$, and $1$ for the Hyper-YOLO-N, Hyper-YOLO-S, Hyper-YOLO-M, and Hyper-YOLO-L, respectively. This means that the actual count of $n$ at each stage of the models is $[3,6,6,6]$ multiplied by the corresponding depth factor for that scale. These specifications ensure that each scale of the HyperYOLO model is equipped with a backbone that is finely tuned for its size and complexity, enabling efficient feature extraction at multiple scales.

\subsection{Neck} 
Compared to the neck design in YOLOv8, the Hyper-YOLO model introduces the HyperC2Net (Hypergraph-Based Cross-level and Cross-position Representation Network) as its neck component, detailed in \cref{fig:neck}. This innovative structure is an embodiment of the proposed HGC-SCS framework, specifically engineered to encapsulate potential high-order correlations existing within the semantic space.

The HyperC2Net is designed to comprehensively fuse cross-level and cross-position information emanating from the backbone network. By leveraging the hypergraph architecture, it effectively captures the complex interdependencies among feature points across different layers and positions. This allows the model to construct a more intricate and enriched representation of the input data, which is particularly useful for identifying and delineating subtle nuances within the images being processed.
In the context of the Hyper-YOLO model's varying scales, the neck plays a critical role in maintaining the consistency of high-order correlation representation. Since the spatial distribution of feature points can significantly differ between models like Hyper-YOLO-N and Hyper-YOLO-L, with the latter typically having a more dispersed distribution, the HyperC2Net adjusts its approach accordingly by employing different distance thresholds for each model scale, as outlined in \cref{app:tab:neck}, to ensure that the network captures the appropriate level of high-order correlations without succumbing to over-smoothing.
The HyperC2Net's ability to dynamically adapt its threshold values based on the model scale and feature point distribution is a testament to its sophisticated design. It strikes a fine balance between the depth of contextual understanding and the need to preserve the sharpness and granularity of the feature space, thereby enhancing the model's overall performance in detecting and classifying objects within varied and complex visual environments.

\begin{table}[!t]
\caption{Detailed configuration of Hyper-YOLO's neck. $C_{in}$ and $C_{out}$ denote the number of input and output channels of HyperConv, respectively. 
$\epsilon$ denotes the predetermined distance threshold for hypergraph construction.
}
\label{app:tab:neck}
\scriptsize
\centering
\begin{tabular}{lcccccccc}
\toprule
& \multicolumn{5}{c}{Channel of Feature} & \multicolumn{3}{c}{HyperConv} \\ 
& $\mB_1$ & $\mB_1$ & $\mB_3$ & $\mB_4$ & $\mB_5$ & $C_{in}$ & $C_{out}$ & $\epsilon$ \\ \midrule
Hyper-YOLO-N & 16 & 32 & 64 & 128 & 256 & 128 & 128 & 6  \\
Hyper-YOLO-S & 32 & 64 & 128 & 256 & 512 & 256 & 256 & 8  \\
Hyper-YOLO-M & 48 & 96 & 192 & 384 & 576 & 384 & 384 & 10  \\
Hyper-YOLO-L & 64 & 128 & 256 & 512 & 512 & 512 & 512 & 10  \\ \bottomrule
\end{tabular}
\end{table}

\begin{figure*}[htbp]
    \centering
    \begin{minipage}[b]{0.49\textwidth}
        \centering
        \includegraphics[width=\textwidth]{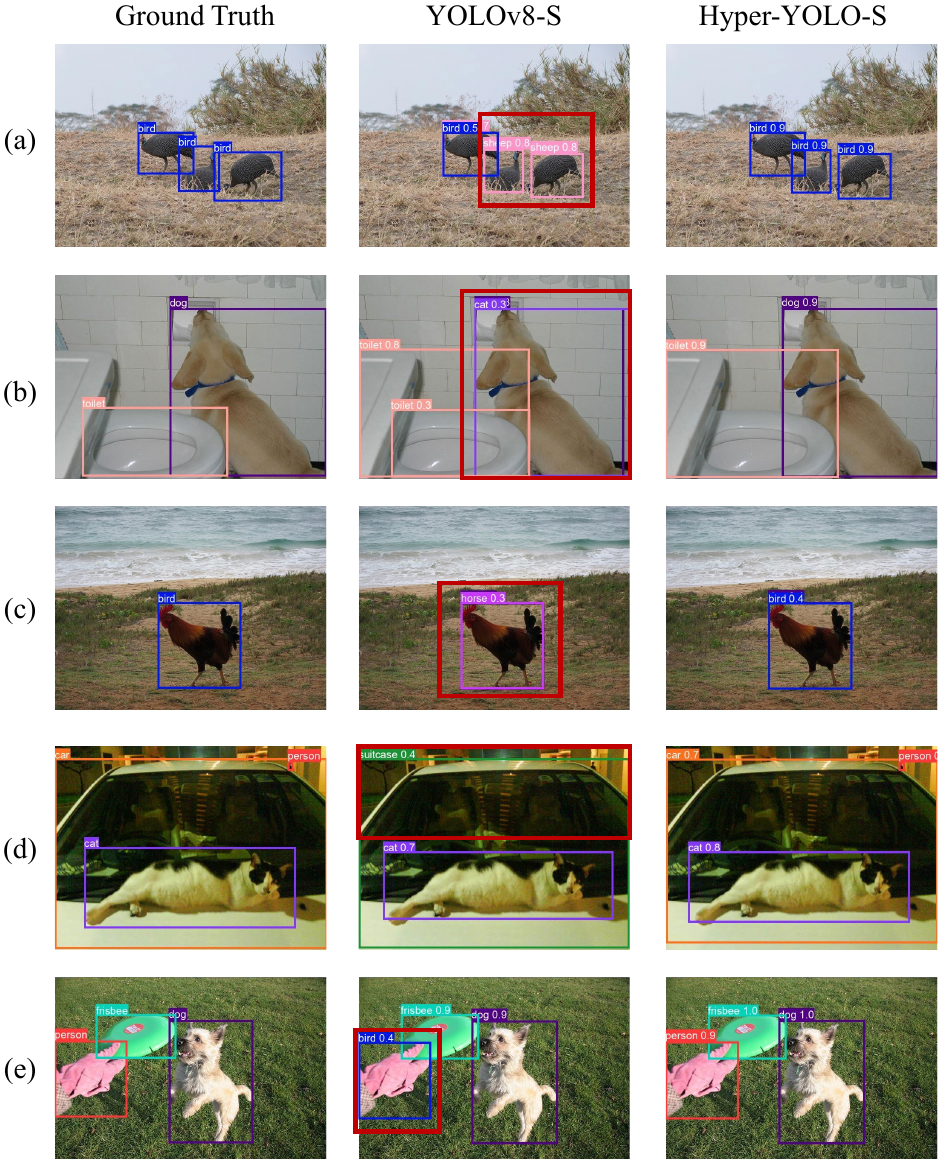}
        \caption{Results comparison of YOLOv8-S and Hyper-YOLO-S in the Object Detection task. Red boxes highlight objects detected with incorrect labels.}
        \label{app:fig:vis-det}
    \end{minipage}
    \hfill
    \begin{minipage}[b]{0.5\textwidth}
        \centering
        \includegraphics[width=\textwidth]{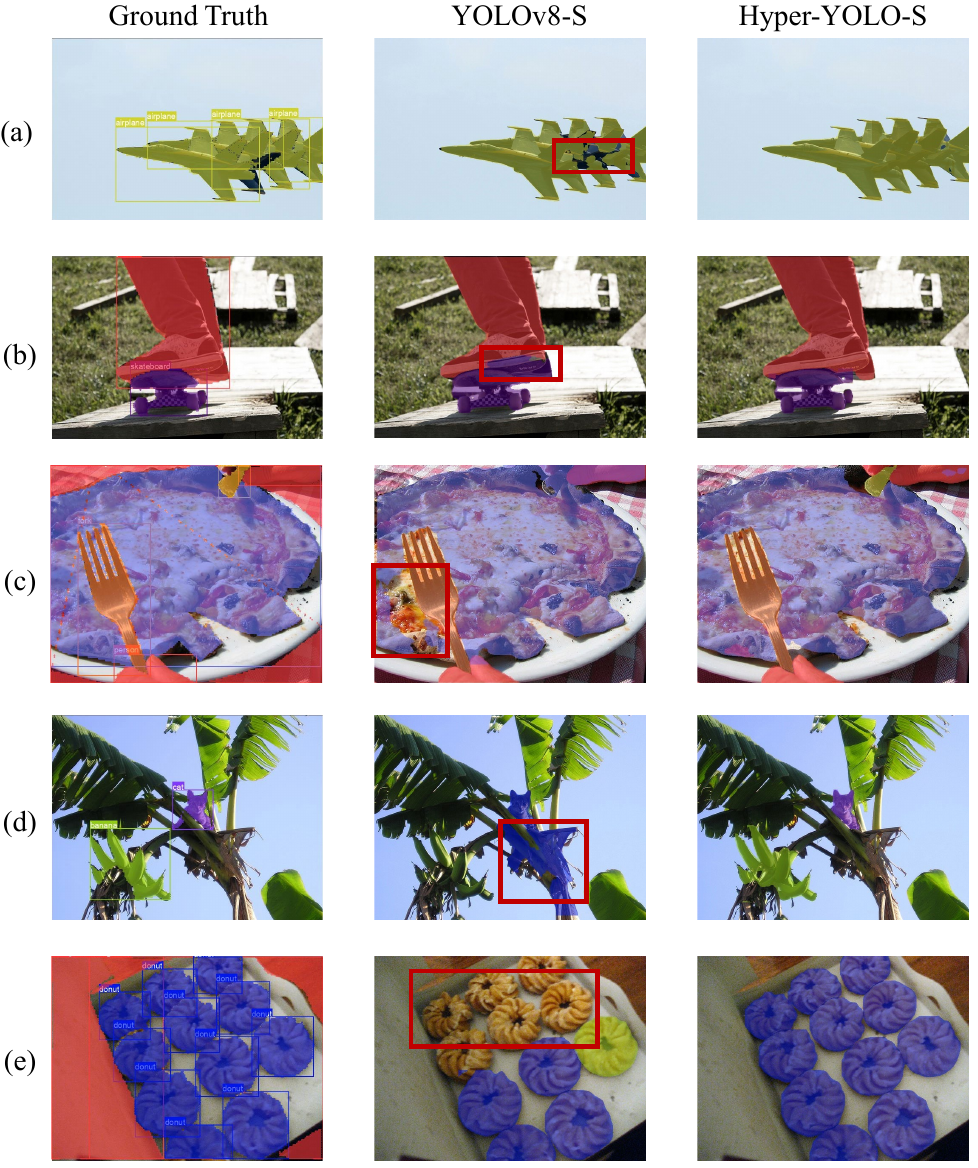}
        \caption{Results comparison of YOLOv8-S and Hyper-YOLO-S in the Instance Segmentation task. Red boxes denote mis-detected or partially segmented. }
        \label{app:fig:vis-seg}
    \end{minipage}
\end{figure*}

\section{Visualizations of Results}
In this section, we further provide visualizations of the Hyper-YOLO on two tasks: object detection and instance segmentation, as shown in \cref{app:fig:vis-det} and \cref{app:fig:vis-seg}, respectively.

\subsection{Object Detection} 
The results depicted in \cref{app:fig:vis-det} illustrate that our Hyper-YOLO model exhibits superior object recognition capabilities, as demonstrated in figures (b) and (c). Moreover, owing to the usage of a hypergraph-based neck in its architecture, Hyper-YOLO possesses a certain degree of class inference ability. This is most evident in figure (a), where Hyper-YOLO is capable of inferring with high confidence that if one bird is detected, the other two entities are also birds. Additionally, as observed in figure (e), it is common for humans to play with dogs using a frisbee. Even though only a glove is visible in the image, our Hyper-YOLO is still able to recognize it as part of a human.

\subsection{Instance Segmentation} 
Results from \cref{app:fig:vis-seg} indicate that, compared to YOLOv8, Hyper-YOLO achieves significant improvements in both categorization and boundary delineation for segmentation tasks. Despite the ground truth annotation in figure (a) not being entirely accurate, our Hyper-YOLO still manages to provide precise boundary segmentation. 
Figures (c), (d) and (e) depict more complex scenes, yet our Hyper-YOLO continues to deliver accurate instance segmentation results, ensuring that not a single cookie is missed.

\section{Training Details of Hyper-YOLO}
\label{app:sec:training}
\begin{table}[!t]
\caption{Detailed configuration of different Hyper-YOLO variations for training.}
\label{app:tab:params}
\centering
\begin{tabular}{l@{\hspace{1cm}}c@{\hspace{0.3cm}}c@{\hspace{0.3cm}}c@{\hspace{0.3cm}}ccccc}
\toprule
Hyperparameter  & $\mN$ & $\mS$ & $\mM$ & $\mL$  \\ \midrule
Epochs                      & 500 & 500 & 500 & 500   \\
Optimizer                   & SGD & SGD & SGD & SGD   \\
lr0                         & 0.01 & 0.01 & 0.01 & 0.01  \\
lrf                         & 0.02 & 0.01 & 0.1 & 0.1  \\
lr decay                    & linear & linear & linear & linear  \\
Momentum                    & 0.937 & 0.937 & 0.937 & 0.937  \\
Weight decay                & 0.0005 & 0.0005 & 0.0005 & 0.0005  \\
Warm up epochs               & 3.0 & 3.0 & 3.0 & 3.0  \\
Warm up momentum             & 0.8 & 0.8 & 0.8 & 0.8  \\
Warm up bias learning rate   & 0.1 & 0.1 & 0.1 & 0.1  \\
Box loss gain               & 7.5 & 7.5 & 7.5 & 7.5  \\
Class loss gain             & 0.5 & 0.5 & 0.5 & 0.5  \\
DFL loss gain               & 1.5 & 1.5 & 1.5 & 1.5  \\
HSV hue augmentation        & 0.015 & 0.015 & 0.015 & 0.015  \\
HSV saturation augmentation & 0.7 & 0.7 & 0.7 & 0.7  \\
HSV value augmentation      & 0.4 & 0.4 & 0.4 & 0.4  \\
Translation augmentation    & 0.1 & 0.1 & 0.1 & 0.1  \\
Scale augmentation          & 0.5 & 0.6 & 0.9 & 0.9  \\
Mosaic augmentation         & 1.0 & 1.0 & 1.0 & 1.0  \\
Mixup augmentation          & 0.0 & 0.0 & 0.1 & 0.1  \\
Copy \& Paste augmentation     & 0.0 & 0.0 & 0.0 & 0.1  \\
Close mosaic epochs         & 10 & 10 & 20 & 20  \\
Hypergraph threshold        & 6 & 8 & 10 & 10 \\ \bottomrule
\end{tabular}
\end{table}


The training protocol for Hyper-YOLO was carefully designed to foster consistency and robustness across varying experiments. Each GPU was allocated a uniform batch size of 20 to maintain a consistent computational environment, utilizing a total of 8 NVIDIA GeForce RTX 4090 GPUs. To assess the learning efficacy and generalization capacity, all variants of Hyper-YOLO, including -N, -S, -M, and -L, were trained from the ground up. The models underwent 500 epochs of training without relying on pre-training from large-scale datasets like ImageNet, thereby avoiding potential biases.
The training hyperparameters were fine-tuned to suit the specific needs of the different sizes of the model. \cref{app:tab:params} summarizes the key hyperparameters for each model scale.

Those core parameters, such as the initial learning rate and weight decay, were uniformly set across all scales to standardize the learning process. The hypergraph threshold, however, was varied according to the model scale and batch size. This threshold was configured with a batch size of 20 per GPU in mind, implying that if the batch size were to change, the threshold would need to be adjusted accordingly. Generally, a larger batch size on a single GPU would necessitate a lower threshold, whereas a larger model scale correlates to a higher threshold.

Most hyperparameters remained consistent across the different model scales; nonetheless, parameters such as the learning rate, scale augmentation, mixup augmentation, copy \& paste augmentation, and the hypergraph threshold were tailored for each model scale. Data augmentation hyperparameters were set in accordance with YOLOv5's configuration, with certain modifications for Hyper-YOLO. For instance, the N and S models employed lower levels of data augmentation, with specific adjustments made for the N model's final learning rate (lrf=0.02) and the S model's scale augmentation (scale=0.6). The M and L models, on the other hand, utilized moderate and high levels of data augmentation, respectively, with both scales having the same setting for close mosaic epochs (20).

It should be emphasized that the hypergraph threshold is set under the premise of a batch size of 20 per GPU. \textit{Alterations to the batch size should be accompanied by corresponding adjustments to the threshold, following the trend that a larger single-GPU batch size should lead to a smaller relative threshold. Similarly, larger model scales require higher thresholds.} Most hyperparameters are consistent across different model scales, with the exception of a few like lrf, scale augmentation, mixup augmentation, copy \& paste augmentation, and hypergraph threshold, which are tailored to the specific scale of the model. Data augmentation parameters are largely based on the YOLOv5 settings, with some values being distinct to accommodate the different requirements of the Hyper-YOLO model.

\section{Details of Speed Test}

The speed benchmarking for our Hyper-YOLO model adopts a two-group approach. The first group comprises models requiring reparameterization, such as YOLOv6-3.0 and Gold-YOLO. The second group includes YOLOv5, YOLOv8, and HyperYOLO. Notably, during the conversion process to ONNX format, the HyperYOLO model encounters issues with the `torch.cdist' function, leading to large tensor sizes that cause errors at batch sizes of 32. 
To address this and ensure accurate speed measurements, we replace the `torch.cdist' function with a custom feature distance function during testing. In addition, we also test the speed of a variant with only an enhanced backbone.

The benchmarking process involves converting the models to ONNX format, followed by conversion to TensorRT engines. The tests are performed twice, under batch sizes of 1 and 32, to assess performance across different operational contexts.
Our test environment is controlled, consisting of Python 3.8.16, Pytorch 2.0.1, CUDA 11.7, cuDNN 8.0.5, TensorRT 8.6.1, and ONNX 1.15.0. All tests are carried out with a fixed input size of $640 \times 640$ pixels.



\vfill

\end{document}